\documentclass{article} 
\usepackage{iclr2027_conference,times}

\usepackage{amsmath,amsfonts,bm}

\def\eqref#1{equation~\ref{#1}}

\def\1{\bm{1}}

\DeclareMathAlphabet{\mathsfit}{\encodingdefault}{\sfdefault}{m}{sl}
\SetMathAlphabet{\mathsfit}{bold}{\encodingdefault}{\sfdefault}{bx}{n}

\usepackage{lipsum}
\usepackage{hyperref}
\usepackage{url}
\usepackage{graphicx}
\usepackage{float}
\usepackage{booktabs}
\usepackage[table]{xcolor}
\usepackage{multirow}
\usepackage{wrapfig}
\usepackage{caption}
\usepackage{subcaption}
\title{Trace4D: Grounded Spatiotemporal Reasoning via 4D State Sequences}

\title{\centering Trace4D: Grounded Spatiotemporal\\Reasoning via 4D State Sequences}

\author{%
\parbox{\dimexpr\textwidth-2\tabcolsep\relax}{\centering
  \vspace*{0.4em}%
  Wooseok Jang\textsuperscript{1} \hspace{2.5em}
  Seungryong Kim\textsuperscript{1} \hspace{2.5em}
  Yan Lu\textsuperscript{2} \hspace{2.5em}
  Lei Ji\textsuperscript{2} \\[0.2em]
  {\normalfont
    \textsuperscript{1}KAIST AI \hspace{2.5em}
    \textsuperscript{2}Microsoft Research Asia \\[0.2em]
    {\small\href{https://cvlab-kaist.github.io/Trace4D}{\texttt{https://cvlab-kaist.github.io/Trace4D}}}
  }
}}

\iclrfinalcopy 
\begin{document}

\maketitle
\lhead{}
\renewcommand{\headrulewidth}{0pt}
\vspace{-2em}                                        
\begin{figure}[H]
    \centering
    \vspace{-10pt}
    \includegraphics[width=0.96\textwidth]{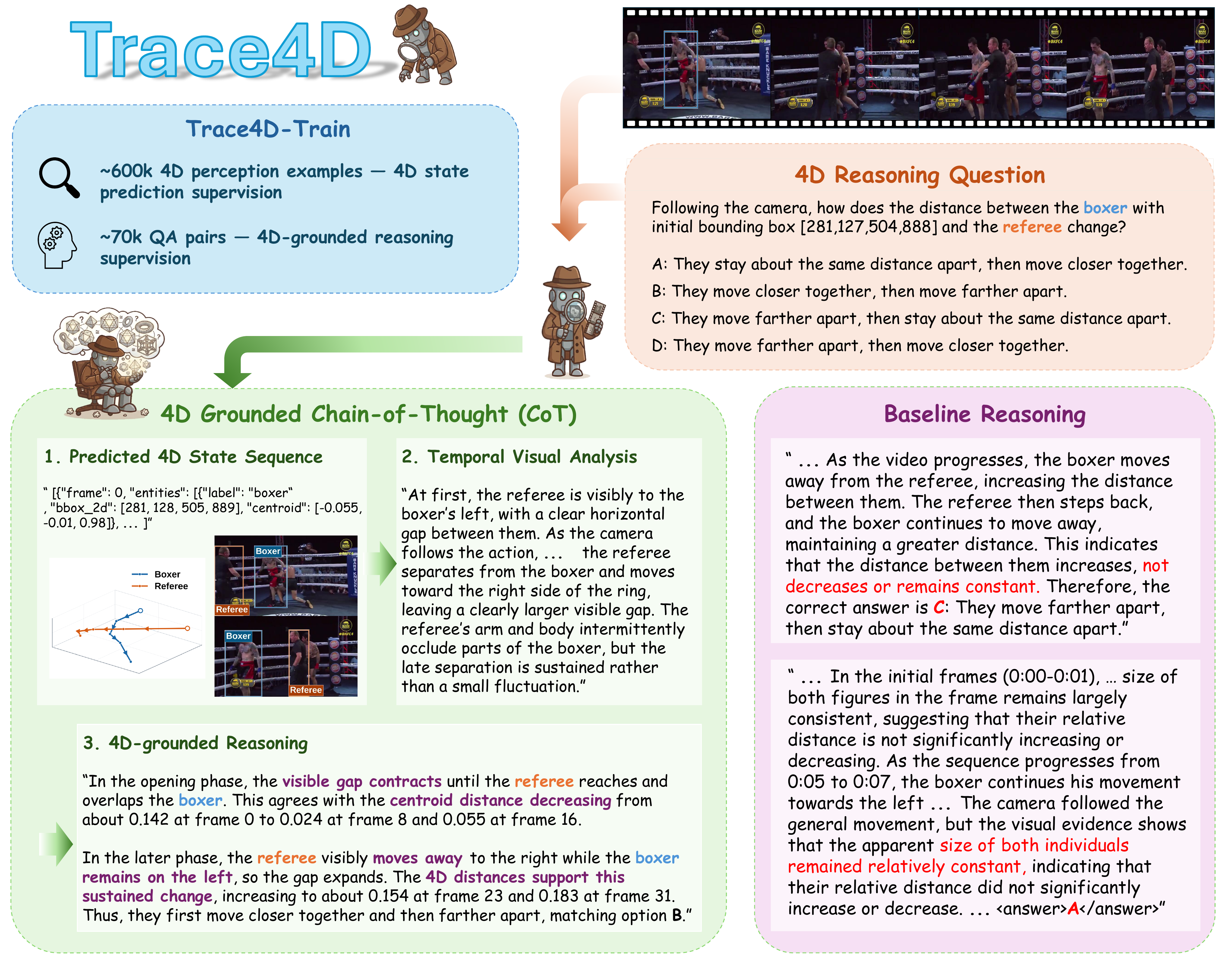}
    \vspace{-10pt}
    \caption{\textbf{Trace4D} connects structured 4D states with temporal visual observations for video reasoning. In the boxer--referee example, centroid-distance changes are matched to the visible approach and subsequent separation, supporting answer B. Baseline reasoning is from Qwen3-VL (top) and MLLM-4D (bottom).}
    \label{fig:teaser}
    \vspace{-15pt}
\end{figure}

\begin{abstract}
Understanding how objects and their spatial relationships evolve in 3D space over time remains challenging for vision-language models (VLMs), particularly under changing viewpoints. Recent approaches incorporate geometric information into reasoning, but explicit supervision for temporally ordered 4D state evolution and its use as intermediate evidence remains limited. We introduce \textbf{Trace4D}, which trains VLMs to predict 4D state sequences and reason over their inter-timestep changes alongside temporal visual observations. Its 4D-grounded chain-of-thought (CoT) explicitly represents 4D states, analyzes temporal visual context, and connects both as evidence for the answer. We construct Trace4D-Train from in-the-wild videos with 600K 4D perception examples and approximately 70K grounded QA examples, and train with joint supervised fine-tuning followed by geometry-aware GRPO. At inference, Trace4D predicts 4D states and reasoning directly from video without reference annotations or an external geometric model. Experiments on Trace4D-Bench and VLM4D show improved QA accuracy over the base VLMs and benefits from reasoning over 4D state changes and temporal visual observations.


\end{abstract}

\section{Introduction}
Understanding how objects and their spatial relationships evolve in 3D space over time is an important capability for video understanding, which we refer to as 4D understanding. Spatial relationships at a single moment provide only a partial account of an event; understanding its development requires relating observations across time. Videos capture only 2D projections of the 3D world, making changes in 3D position and spatial relationships difficult to infer directly. This makes 4D reasoning challenging for vision-language models (VLMs)~\citep{yang2025thinking,zhou2025vlm4d}.

To address this challenge, recent approaches have equipped VLMs with geometric knowledge in different ways. Some incorporate pretrained geometric features or outputs~\citep{liu2026geoalign,zhou2026learning,gao2026dyntrace}, providing spatial references that complement the VLM's visual representations. Others transfer 4D knowledge through QA supervision~\citep{zhou2026learning,cho20264dp}, sometimes complemented by perception and tracking tasks. Recent approaches also extend chain-of-thought (CoT) reasoning~\citep{wei2022chain} with geometric states, relations, and temporal visual evidence~\citep{yin2026mllm,wang2025video}. However, explicit supervision for predicting a temporally ordered 4D state sequence remains limited. Such a sequence makes inter-timestep changes explicit, allowing the model to compare states across time and use these changes as intermediate evidence for reasoning.

Grounded reasoning methods link intermediate reasoning steps to identifiable visual evidence. Image-based approaches localize relevant object regions and connect them to reasoning steps~\citep{wu2026grounded,wang2026traceable}, while video methods extend this principle to specific objects and timestamps~\citep{han2025videoespresso,meng2025open}. For 4D understanding, the relevant evidence includes both how 3D states change over time and the visual context in which those changes occur. A 4D state sequence provides a basis for comparing states across time, while temporal visual observations describe actions, interactions, and occlusions that geometric states alone do not fully capture. This motivates grounded temporal reasoning that connects state changes to their visual context and uses both sources of evidence to support the final answer.

We introduce \textbf{Trace4D}, which learns to predict a 4D state sequence as an explicit intermediate representation for video reasoning. The sequence describes object and, when required, camera states across annotated timesteps. Our 4D-grounded CoT uses this sequence to interpret state changes alongside temporal visual observations and derive an answer (Figure~\ref{fig:teaser}). We construct Trace4D-Train with 600K 4D perception examples and approximately 70K grounded QA examples from in-the-wild videos to supervise state prediction and its use in reasoning. Joint supervised fine-tuning (SFT) learns both tasks together. Geometry-aware GRPO further rewards agreement with reference positions and inter-timestep displacements. At inference, the model generates its own state sequence and reasoning from the video and QA prompt without an external geometric model. Trace4D achieves the highest overall accuracy among evaluated models on our held-out \textbf{Trace4D-Bench} and improves overall accuracy over its base VLMs on VLM4D.

Our contributions are summarized as follows:
\begin{itemize}
    \item We introduce \textbf{Trace4D}, which predicts temporally ordered 4D state sequences as explicit intermediate representations and grounds reasoning over their changes in temporal visual observations.

    \item We construct \textbf{Trace4D-Train} with 600K 4D perception and approximately 70K grounded QA examples from in-the-wild videos, supervising state prediction and its use in reasoning.

    \item Ablations show that reasoning over 4D state changes and geometry-aware GRPO improve performance, while temporal visual analysis provides complementary evidence for spatiotemporal reasoning.
\end{itemize}

\section{Related Works}

\paragraph{VLMs for 3D and 4D Understanding.}

General-purpose VLMs support video QA~\citep{zhang2024llava,Bai2025Qwen25VLTR,bai2025qwen3}, but evaluations of spatial and dynamic scene understanding reveal persistent challenges~\citep{yang2025thinking,zhou2025vlm4d,zhou2026learning}. To strengthen these capabilities, geometry-derived QA provides supervision for spatial relations in images~\citep{chen2024spatialvlm,cheng2024spatialrgpt} and kinematic changes along object trajectories~\citep{ko2025st}. Multitask supervision further covers depth, correspondence, and camera/object motion~\citep{xu2026multi}, as well as 4D QA, tracking, and camera-pose tasks~\citep{zhou2026learning,cho20264dp,yang2026cambrian}. Geometric knowledge can also be supplied through pretrained features~\citep{liu2026geoalign,miao2026geoweaver} or constructed as explicit evidence at inference time~\citep{gao2026dyntrace}. Several works also use geometry as an explicit intermediate representation for reasoning: 3D locations and orientations enable spatial computation~\citep{ma2026spatialreasoner}, while generated scene-layout maps guide reasoning through rewards for spatial accuracy~\citep{ouyang2025spacer}. For video reasoning, relation graphs make inter-object geometry explicit~\citep{wang2025video}. MLLM-4D~\citep{yin2026mllm} predicts explicit 3D states at query-relevant start and end frames, analyzes the intervening visual progression, and rewards coordinate accuracy at those anchors. In contrast, Trace4D predicts a multi-timestep 4D state sequence as an explicit intermediate representation and uses its inter-timestep changes as reasoning evidence alongside temporal visual observations.

\paragraph{Grounded reasoning for VLMs.}

Chain-of-thought (CoT) prompting elicits intermediate reasoning before an answer~\citep{wei2022chain}. Early studies extend rationale supervision to multimodal QA~\citep{lu2022learn,zhang2023multimodal}, while LLaVA-CoT~\citep{xu2025llava} separates visual interpretation and reasoning into stages. Grounded reasoning links these steps to identifiable visual evidence. At the image level, VisCoT~\citep{shao2024visual} selects informative regions, while VoCoT~\citep{li2025vocot} incorporates localized object information into the reasoning sequence. This connection can be learned through stepwise grounding supervision~\citep{wu2026grounded} or joint localization and reasoning rewards~\citep{wang2026traceable}, with further visual feedback obtained by revisiting selected or zoomed regions~\citep{wang2026vgr,sarch2026grounded}. For video, evidence must also be associated with temporal locations: VideoEspresso~\citep{han2025videoespresso} annotates frames, objects, and textual evidence, while Open-o3-Video~\citep{meng2025open} links reasoning to timestamps, objects, and bounding boxes. Related settings ground motion reasoning through segmentation outputs~\citep{deng2025motion} or connect 3D scene entities to reasoning steps and subproblems~\citep{chen2026reasoning,linghu2026scenecot}. Existing video-grounding approaches localize supporting evidence in space and time. Trace4D additionally supervises reasoning over predicted 4D state changes alongside the corresponding temporal visual observations.


\section{Methodology}
\label{sec:methodology}

We present Trace4D, a pipeline for learning to predict 4D state sequences and use their changes together with temporal visual observations for video reasoning. We first construct reference state sequences and QA pairs from in-the-wild videos (Section~\ref{sec:data_generation}). A multimodal teacher then produces 4D-grounded CoT that relates these state changes to temporal visual observations and uses both as evidence for the answer (Section~\ref{sec:cot_generation}). Finally, we jointly train the VLM on state prediction and grounded QA, followed by geometry-aware GRPO using answer correctness and geometric rewards (Section~\ref{sec:stagewise_training}).

\suppressfloats[t]
\begin{figure}[!t]
    \centering
    \includegraphics[width=\textwidth,height=\textwidth,keepaspectratio]{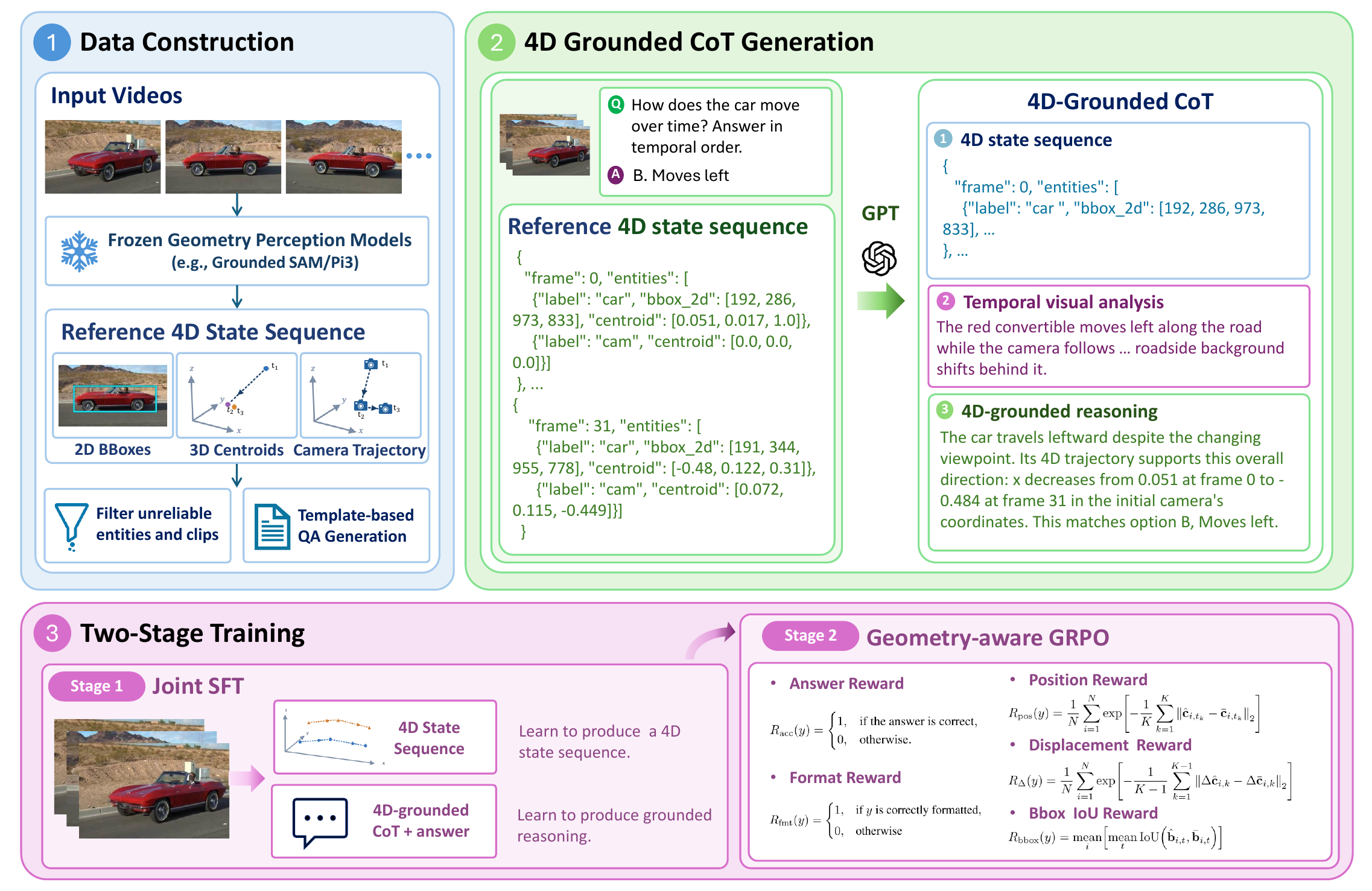}
    \caption{Overview of \textbf{Trace4D}. Reference 4D states support template-based QA construction and teacher-generated CoT that connects state changes with temporal visual observations. Joint SFT learns state prediction and grounded QA together, followed by geometry-aware GRPO using answer, format, position, displacement, and bounding-box rewards.}
    \label{fig:main}
\end{figure}

\subsection{Dataset Generation Pipeline}
\label{sec:data_generation}

Our data pipeline uses the geometric perception components and selected answer-derivation rules of DSR-Suite~\citep{zhou2026learning}. We further filter the resulting tracks and adapt the question and option templates to our clip-level setting.

\paragraph{Video and entity curation.}
We curate source videos from Koala-36M~\citep{wang2025koala} using its motion and video-training-suitability metadata to identify videos with observable scene or object dynamics. We identify candidate agents and foreground objects from the captions and use the Grounded SAM pipeline~\citep{ren2024grounded}, with Grounding DINO~\citep{liu2024grounding} for localization and SAM~2~\citep{ravi2025sam} for mask tracking across frames.

\begin{wrapfigure}{r}{0.48\textwidth}
    \vspace{-\intextsep}
    \centering
    \includegraphics[width=\linewidth]{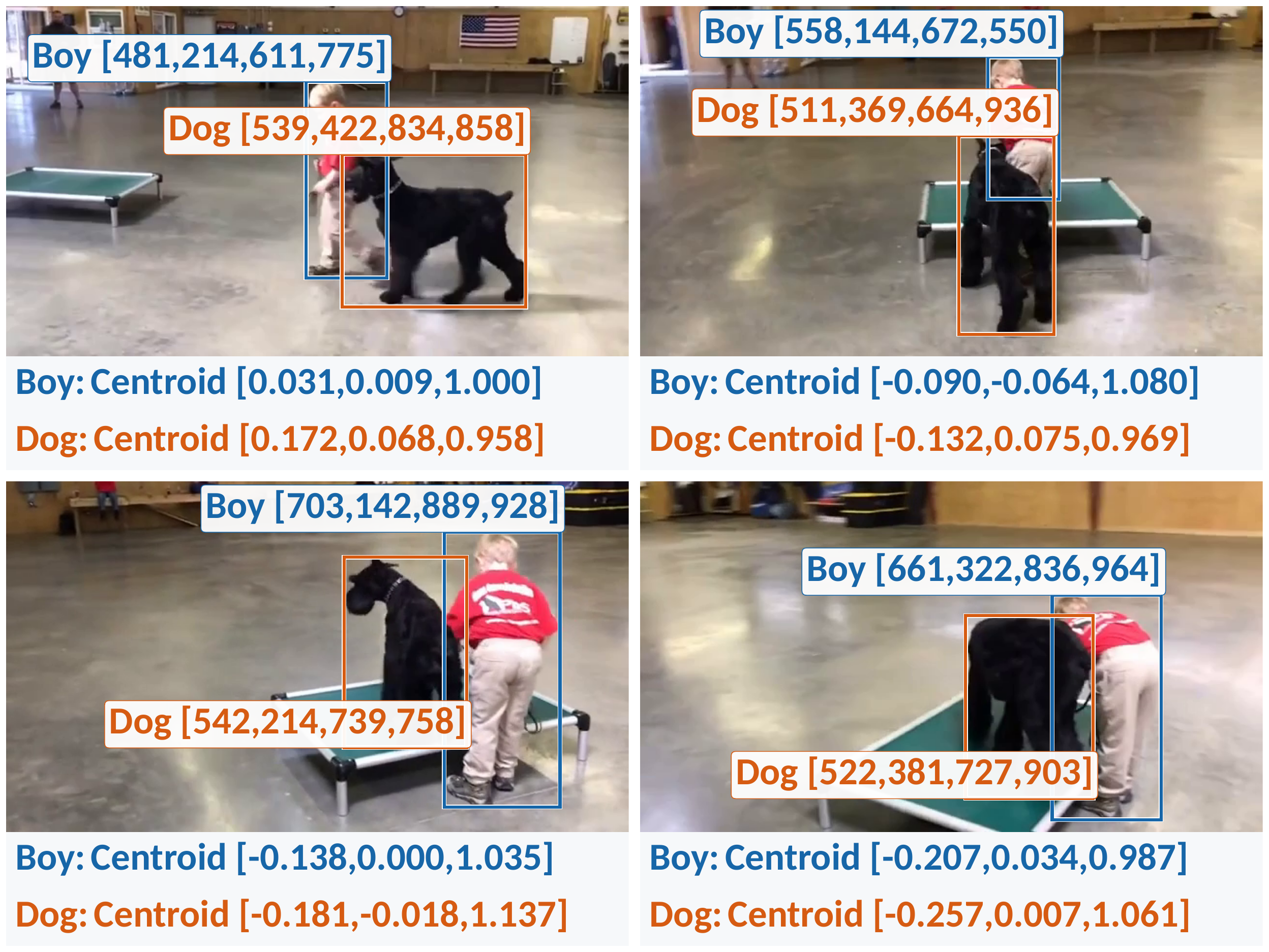}
    \vspace{-20pt}
    \caption{Example of 4D reference annotations. Colored boxes identify the same two entities across four frames; labels report their bounding-box coordinates and 3D centroids.}
    \label{fig:annotation_example}
    \vspace{-25pt}
\end{wrapfigure}

\paragraph{Reference 4D state construction.}
Pi3~\citep{wang2026pi} estimates 3D point maps and camera poses. We obtain object centroids by averaging reconstructed points within each tracked mask and derive camera positions from the estimated poses. These estimates form our \textit{reference 4D states} (Figure~\ref{fig:annotation_example}), distinct from student predictions.

For a clip represented by $T$ sampled frames, $V=(I_t)_{t=0}^{T-1}$, let $\mathcal{K}=\{t_1,\ldots,t_K\}\subseteq\{0,\ldots,T-1\}$ denote its $K$ annotated frame indices, ordered as $t_1<\cdots<t_K$. Each state $(t,\mathcal{E}_t)$ describes the target entities at timestep $t$, and the 4D state sequence is
\begin{equation}
\begin{aligned}
\mathcal{P} &= \bigl[(t,\mathcal{E}_t)\bigr]_{t\in\mathcal{K}},\\
\mathcal{E}_t &= \bigl\{(\ell_i,\mathbf{b}_{i,t},\mathbf{c}_{i,t})\bigr\}_{i\in\mathcal{O}_t}.
\end{aligned}
\label{eq:perceptual_state}
\end{equation}

Here, $\mathcal{O}_t$ indexes the annotated entities and $\ell_i$ is the label of entity $i$. The index $i$ refers to the same entity across timesteps. For an object, $\mathbf{b}_{i,t}\in\mathbb{R}^{4}$ is its normalized 2D box and $\mathbf{c}_{i,t}\in\mathbb{R}^{3}$ its normalized 3D centroid. Camera entries use $\ell_i=\texttt{cam}$, $\mathbf{c}_{i,t}$ for position, and $\mathbf{b}_{i,t}=\emptyset$ to denote an omitted box.

Appendix~\ref{app:state_serialization} defines the serialized fields, frame indexing, and coordinate normalization.

\paragraph{Quality filtering and clip selection.}
We filter the resulting object tracks using bounding-box size checks and a multimodal verifier that rejects empty, background, or incorrectly labeled target regions. We then select temporal windows with identifiable targets and valid required bounding boxes and 3D centroids, ensuring that their annotations are indexed to the sampled clip timeline. Unreadable clips that cannot support the required frame sampling are excluded. The retained states serve as references for perception and QA supervision.

\paragraph{Template-based QA construction.}
We construct QA pairs under two coordinate-frame conventions. \textit{Relative questions} express spatial relations and their changes in the camera frame at each timestep, while \textit{Absolute questions} express them in a fixed anchor-camera frame for comparison over time. Here, \textit{absolute} refers to a fixed reference frame rather than metric world coordinates.

For each question, we transform the reference 4D states into the required coordinate frame, yielding the task-specific reference sequence $\bar{\mathcal{P}}$. An overbar denotes a training target. We then derive the answer from the resulting temporal state changes using selected geometric rules adopted from DSR-Suite~\citep{zhou2026learning}. We revise the question and option templates for our clip-level setting and additionally construct motion questions within the Absolute group for object and camera translation over time. For task-specific distractors, we give priority to candidates with the same number of temporal phases as the reference answer to reduce answer-length cues. Appendix~\ref{app:qa_templates} gives question templates, answer criteria, and distractor rules. Each resulting QA pair is retained with its source clip and aligned reference 4D states for subsequent CoT generation.
\subsection{4D-Grounded CoT Generation}
\label{sec:cot_generation}

We next augment the QA pairs with 4D-grounded CoT annotations, keeping the questions, answer options, and reference answers fixed. Let $q$ denote the complete QA prompt, including the question, answer options, and task instruction. A multimodal teacher receives $V$, $q$, the reference 4D state sequence $\bar{\mathcal{P}}$, and the reference option label $a_{\mathrm{ref}}\in\{\mathrm{A},\mathrm{B},\mathrm{C},\mathrm{D}\}$ obtained during QA construction. The teacher constructs the following three-part CoT to connect visual observations with state changes in support of the reference answer.

\textit{Step 1: Structured 4D state sequence.}
The teacher copies the task-specific reference states unchanged into the \texttt{<4d>} block at every annotated timestep in $\mathcal{K}$, providing the student's state-prediction target.

\textit{Step 2: Temporal visual analysis.}
The \texttt{<4d>} block represents geometry but does not fully capture visual cues such as actions, interactions, and occlusion. The \texttt{<visual>} block therefore records question-relevant observations across temporal phases.

\textit{Step 3: 4D-grounded reasoning.}
Question-type-specific instructions require the teacher to connect a temporal visual observation with the relevant 4D state change for the same target, interval, and coordinate frame, then explain how this correspondence supports the answer. When the two sources differ, the teacher is instructed to re-examine relevant viewpoint or occlusion effects without forcing agreement.

Let $s$, $o$, $r$, and $a$ denote the token spans for the structured 4D state sequence, temporal visual analysis, 4D-grounded reasoning, and final answer. Each span includes its associated formatting tokens, so their concatenation forms the complete response $y=[s;o;r;a]$.

The teacher-generated target is $\bar{y}=[\bar{s};\bar{o};\bar{r};\bar{a}]$, where $\bar{s}$ serializes $\bar{\mathcal{P}}$ and $\bar{a}$ encodes $a_{\mathrm{ref}}$. It is paired with $(V,q)$ for student training; the reference state sequence and correct-option label are not added as extra student inputs. We retain traces that preserve the reference states and answer and satisfy the required response structure. Appendix~\ref{app:cot_generation_settings} describes the teacher inputs and provides an example generation prompt.

\begin{figure*}[t]
    \centering
    
    \begin{subfigure}{0.32\linewidth}
        \centering
        \includegraphics[width=\linewidth]{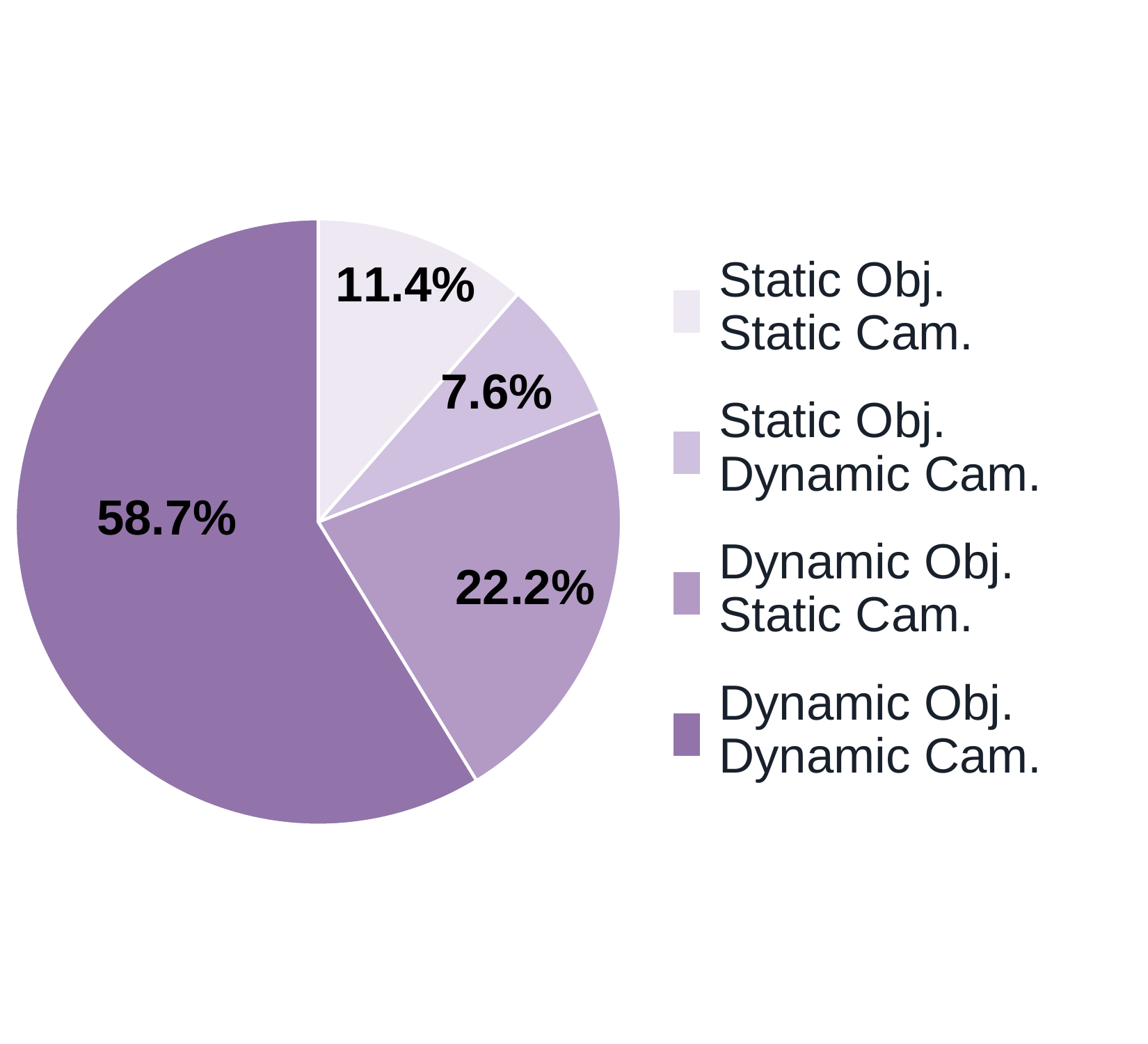} 
        \caption{Dynamics distribution}
        \label{fig:stat_a}
    \end{subfigure}\hfill
    \begin{subfigure}{0.37\linewidth}
        \centering
        \includegraphics[width=\linewidth]{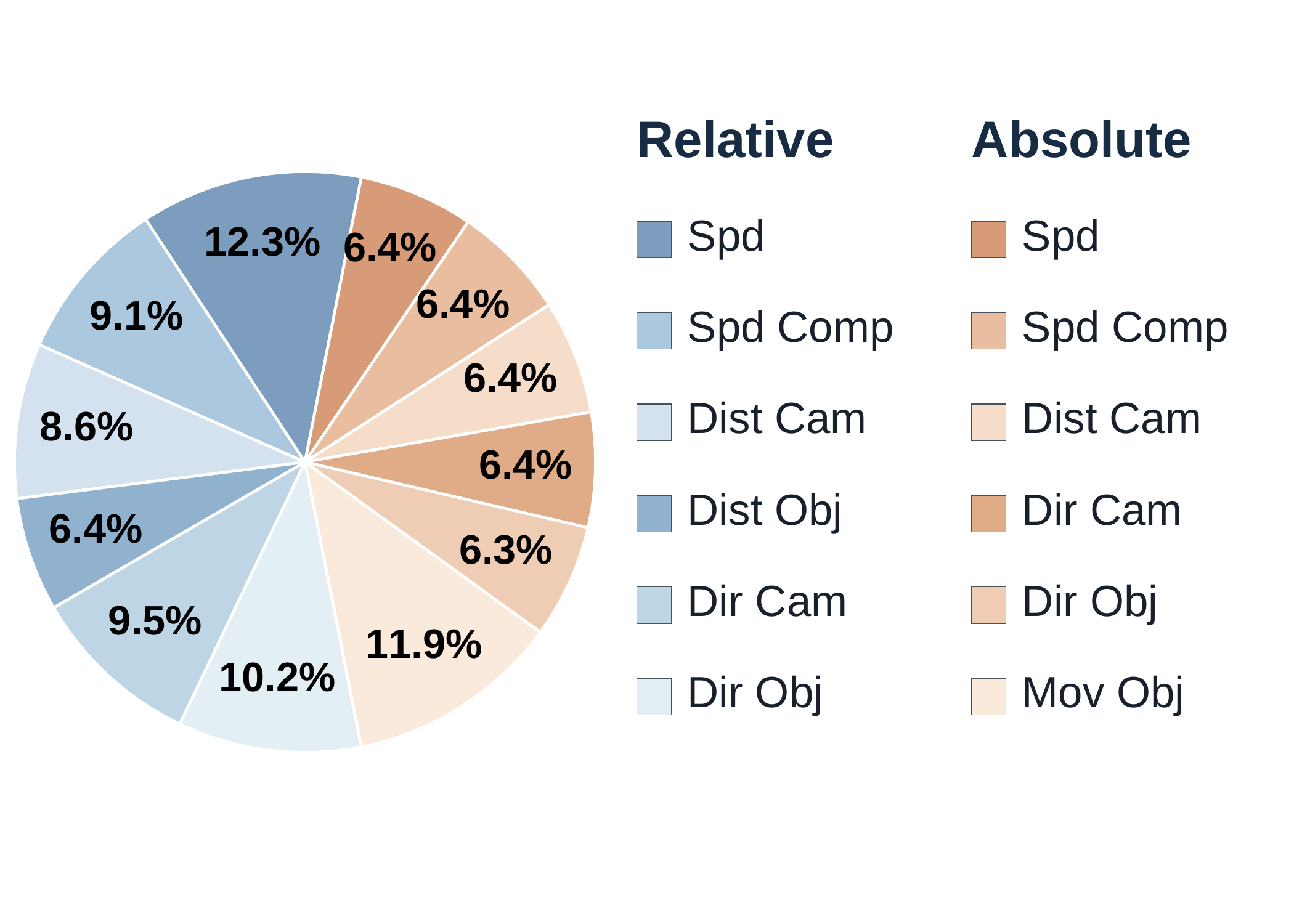}
        \caption{Question-type distribution}
        \label{fig:stat_b}
    \end{subfigure}\hfill
    \begin{subfigure}{0.28\linewidth}
        \centering
        \includegraphics[width=\linewidth]{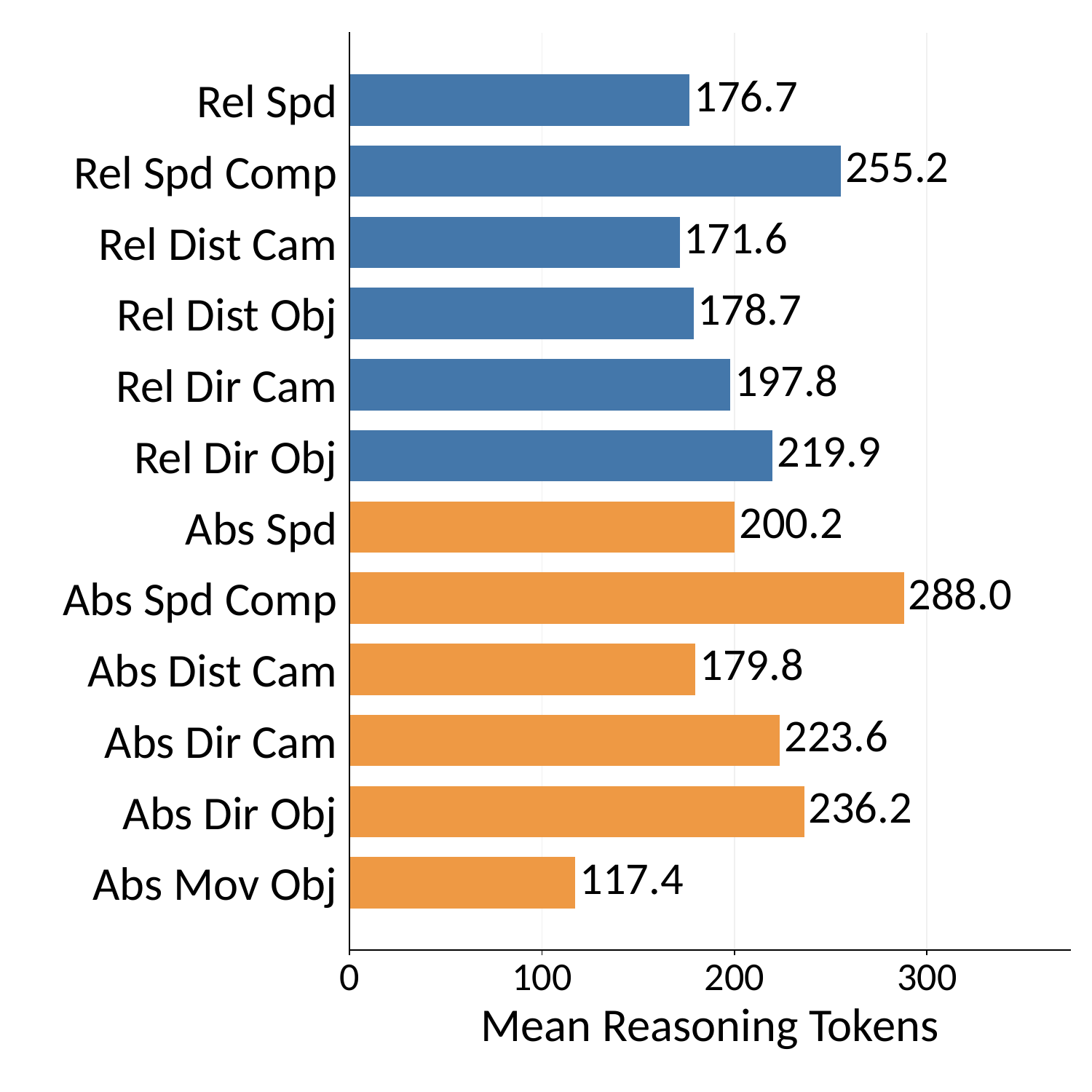}
        \caption{Mean reasoning length}
        \label{fig:stat_c}
    \end{subfigure}
    
    \caption{Dataset statistics. (a) Joint distribution of object and camera dynamics. (b) Question-type proportions for relative and absolute QA. (c) Mean reasoning token counts by question type.}
    \label{fig:data_statistics}
\end{figure*}

\paragraph{Training--evaluation splits and statistics.}
Our data construction yields 600K \textit{4D perception examples} from approximately 362K unique clips and 69,331 \textit{grounded QA examples} (approximately 70K) from approximately 64K unique clips. A clip may contribute to either form of supervision or both. We call the training corpus \textbf{Trace4D-Train}. Splits are defined at the parent source-video level, with held-out sources excluded from all training, including GRPO and ablations. \textbf{Trace4D-Bench} is the held-out evaluation set of 1,200 QA pairs: 600 Relative and 600 Absolute questions, with 100 questions per type.

Figure~\ref{fig:data_statistics} summarizes object--camera dynamics, question-type coverage, and reasoning length in the constructed dataset. Joint object and camera motion forms the largest dynamics category, while speed-comparison questions have the longest mean reasoning traces in both coordinate settings.

\subsection{Two-Stage Training}
\label{sec:stagewise_training}

We train Trace4D in two stages: joint supervised fine-tuning (SFT) on 4D perception and grounded QA, followed by geometry-aware GRPO.

\paragraph{Stage 1: Joint SFT.}
Starting from the base VLM, we jointly train on Trace4D-Train's 600K perception and approximately 70K grounded QA examples. A perception example provides a clip $V$ and an instruction $u$ specifying the targets and coordinate convention. The target serializes the reference 4D states $\bar{\mathcal{P}}$ within \texttt{<4d>}, without visual analysis or an answer. A grounded QA example provides $(V,q)$ and supervises the full response $\bar{y}$. Appendix~\ref{app:shared_prompts} shows both task instructions; Section~\ref{sec:ablation_analysis} compares joint SFT with separate perception pretraining.

\paragraph{Stage 2: Geometry-aware GRPO.}
Starting from the joint-SFT checkpoint, we apply Group Relative Policy Optimization (GRPO)~\citep{shao2024deepseekmath} on approximately 7K QA examples to reward answer correctness and geometric consistency. Reference states and answer labels are used only for reward computation. The position reward $R_{\mathrm{pos}}$ rewards position accuracy. The displacement reward $R_{\Delta}$ compares coordinate-change vectors between consecutive annotated timesteps, accounting for direction and magnitude. Both rewards are computed from predicted states in the generated \texttt{<4d>} block. For $N$ entities, including the camera when required, matched across $K\geq2$ valid annotated timesteps, these rewards are
\begin{equation}
\begin{aligned}
R_{\mathrm{pos}}(y)
&=\frac{1}{N}\sum_{i=1}^{N}\exp\!\left[
-\frac{1}{\tau_{\mathrm{pos}}K}
\sum_{k=1}^{K}
\left\|\hat{\mathbf{c}}_{i,t_k}-\bar{\mathbf{c}}_{i,t_k}\right\|_2
\right],\\
R_{\Delta}(y)
&=\frac{1}{N}\sum_{i=1}^{N}\exp\!\left[
-\frac{1}{\tau_{\Delta}(K-1)}
\sum_{k=1}^{K-1}
\left\|\Delta\hat{\mathbf{c}}_{i,k}-\Delta\bar{\mathbf{c}}_{i,k}\right\|_2
\right],
\end{aligned}
\label{eq:geometry_rewards}
\end{equation}
where $\hat{\mathbf{c}}$ and $\bar{\mathbf{c}}$ are predicted and reference positions under the same coordinate and normalization conventions, $\Delta\mathbf{c}_{i,k}=\mathbf{c}_{i,t_{k+1}}-\mathbf{c}_{i,t_k}$, and $\tau_{\mathrm{pos}}=\tau_{\Delta}=1$. Absolute displacements use the fixed anchor-camera frame; relative displacements use camera-relative coordinates.

The binary rewards $R_{\mathrm{acc}}$ and $R_{\mathrm{fmt}}$ score answer correctness and response format; $R_{\mathrm{bbox}}$ averages bounding-box IoU over matched objects. The combined reward is
\begin{equation}
R(y)=w_{\mathrm{acc}}R_{\mathrm{acc}}(y)
     +w_{\mathrm{fmt}}R_{\mathrm{fmt}}(y)
     +w_{\mathrm{pos}}R_{\mathrm{pos}}(y)
     +w_{\Delta}R_{\Delta}(y)
     +w_{\mathrm{bbox}}R_{\mathrm{bbox}}(y),
\label{eq:composite_reward}
\end{equation}
where the weights sum to one. Appendix~\ref{app:grpo_details} specifies the reward weights, format criteria, and optimization settings.

\paragraph{Inference.}
Given $(V,q)$, Trace4D generates $y=[s;o;r;a]$ autoregressively. The predicted 4D state sequence remains in the context for subsequent temporal visual analysis and 4D-grounded reasoning. Reference 4D states and answer labels are not supplied at inference, and no external geometric model is invoked. Appendix~\ref{app:reasoning_examples} shows examples of the generated responses.

\section{Experiments}

We evaluate whether Trace4D can use predicted 4D state changes and temporal visual observations to answer spatiotemporal questions. We evaluate the trained model on our held-out Trace4D-Bench and examine transfer to the external VLM4D benchmark. Our ablations study the effects of state-change reasoning, temporal visual evidence, joint SFT, and GRPO. We further analyze the relationship between 4D perception quality and QA accuracy. Detailed GRPO reward ablations are provided in Appendix~\ref{app:grpo_ablation}.

\subsection{Experimental Settings}

We adapt Qwen3-VL-4B-Instruct and Qwen3-VL-8B-Instruct~\citep{bai2025qwen3} using LoRA~\citep{hu2021lora} on the language model and multimodal projector, while keeping the vision backbone frozen. Trace4D results use joint SFT followed by geometry-aware GRPO. We evaluate multiple-choice QA accuracy on Trace4D-Bench and VLM4D. All ablations use the 4B model; CoT ablations use matched QA adaptation without GRPO, while the training-strategy ablation compares different SFT recipes and their GRPO continuation. Further details on data mixtures, optimization, task instructions, and decoding settings are provided in Appendix~\ref{app:training_details}.

\begin{table*}[t]
\centering
\setlength{\tabcolsep}{4pt}
\resizebox{\textwidth}{!}{
\begin{tabular}{l|cccccc|>{\columncolor{gray!10}}c|cccccc|>{\columncolor{gray!10}}c|>{\columncolor{gray!10}}c}
\toprule

\multirow{2}{*}{\textbf{Models}} &
\multicolumn{7}{c|}{\textbf{Relative Spatial Relations}} &
\multicolumn{7}{c|}{\textbf{Absolute Spatial Relations}} &
\multicolumn{1}{c}{\multirow{2}{*}{\textbf{Overall}}} \\

\cmidrule(lr){2-8}
\cmidrule(lr){9-15}

&
Spd &
Spd Comp &
Dist$_{\mathrm{Cam}}$ &
Dist$_{\mathrm{Obj}}$ &
Dir$_{\mathrm{Cam}}$ &
Dir$_{\mathrm{Obj}}$ &
\textbf{Avg.} &
Spd &
Spd Comp &
Dist$_{\mathrm{Cam}}$ &
Dir$_{\mathrm{Cam}}$ &
Dir$_{\mathrm{Obj}}$ &
Motion &
\textbf{Avg.} &
\multicolumn{1}{c}{} \\

\midrule

\rowcolor{gray!20}
\multicolumn{16}{c}{\textit{Proprietary Models}} \\

GPT-4o~\citep{hurst2024gpt}
& 33.0 & 32.0 & 21.0 & 13.0 & 47.0 & 32.0 & 29.67
& 31.0 & 32.0 & 36.0 & 38.0 & 35.0 & 54.0 & 37.67
& 33.67 \\

GPT-5~\citep{singh2025openai}
& 32.0 & 37.0 & 58.0 & 47.0 & 68.0 & 54.0 & 49.33
& 35.0 & 36.0 & 50.0 & 40.0 & 43.0 & 60.0 & 44.00
& 46.67 \\

\midrule

\rowcolor{gray!20}
\multicolumn{16}{c}{\textit{General VLMs}} \\

Qwen2.5-VL-3B~\citep{Bai2025Qwen25VLTR}
& 40.0 & 47.0 & 25.0 & 35.0 & 21.0 & 22.0 & 31.67
& 42.0 & 49.0 & 31.0 & 24.0 & 29.0 & 33.0 & 34.67
& 33.17 \\

Qwen3-VL-4B~\citep{bai2025qwen3}
& 26.0 & 38.0 & 30.0 & 41.0 & 41.0 & 53.0 & 38.17
& 29.0 & 33.0 & 41.0 & 38.0 & 47.0 & 50.0 & 39.67
& 38.92 \\

Qwen2.5-VL-7B
& 39.0 & 48.0 & 32.0 & 26.0 & 40.0 & 32.0 & 36.17
& 33.0 & 40.0 & 35.0 & 49.0 & 26.0 & 55.0 & 39.67
& 37.92 \\

Qwen3-VL-8B
& 32.0 & 43.0 & 36.0 & 31.0 & 55.0 & 44.0 & 40.17
& 37.0 & 41.0 & 31.0 & 49.0 & 39.0 & 55.0 & 42.00
& 41.08 \\

InternVL2.5-4B~\citep{chen2024expanding}
& 27.0 & 41.0 & 21.0 & 30.0 & 35.0 & 46.0 & 33.33
& 39.0 & 41.0 & 24.0 & 37.0 & 40.0 & 36.0 & 36.17
& 34.75 \\

InternVL3.5-4B~\citep{wang2025internvl3}
& 41.0 & 43.0 & 30.0 & 43.0 & 40.0 & 39.0 & 39.33
& 42.0 & 36.0 & 37.0 & 41.0 & 42.0 & 46.0 & 40.67
& 40.00 \\

InternVL2.5-8B
& 41.0 & 38.0 & 43.0 & 38.0 & 49.0 & 37.0 & 41.00
& 39.0 & 37.0 & 28.0 & 50.0 & 37.0 & 42.0 & 38.83
& 39.92 \\

InternVL3.5-8B
& 34.0 & 45.0 & 33.0 & 32.0 & 33.0 & 48.0 & 37.50
& 37.0 & 39.0 & 28.0 & 50.0 & 37.0 & 55.0 & 41.00
& 39.25 \\

LLaVA-OneVision1.5-4B~\citep{an2025llava}
& 30.0 & 43.0 & 26.0 & 29.0 & 34.0 & 42.0 & 34.00
& 22.0 & 35.0 & 29.0 & 38.0 & 54.0 & 43.0 & 36.83
& 35.42 \\

LLaVA-OneVision1.5-8B
& 25.0 & 38.0 & 20.0 & 30.0 & 31.0 & 48.0 & 32.00
& 21.0 & 32.0 & 26.0 & 34.0 & 41.0 & 52.0 & 34.33
& 33.17 \\


\midrule

\rowcolor{gray!20}
\multicolumn{16}{c}{\textit{3D/4D Specialized Models}} \\

VG-LLM-8B~\citep{zheng2026learning}
& 39.0 & 45.0 & 36.0 & 43.0 & 87.0 & 18.0 & 44.67
& 44.0 & 47.0 & 38.0 & 91.0 & 25.0 & 44.0 & 48.17
& 46.42 \\

VLM-3R-7B~\citep{fan2026vlm}
& 40.0 & 45.0 & 27.0 & 38.0 & 32.0 & 21.0 & 33.83
& 40.0 & 46.0 & 32.0 & 44.0 & 31.0 & 51.0 & 40.67
& 37.25 \\

MLLM-4D-8B~\citep{yin2026mllm}
& 35.0 & 43.0 & 51.0 & 46.0 & 69.0 & 41.0 & 47.50
& 34.0 & 50.0 & 41.0 & 62.0 & 31.0 & 61.0 & 46.50
& 47.00 \\

GSM-7B~\citep{zhou2026learning}
& 43.0 & 45.0 & 33.0 & 34.0 & 74.0 & 28.0 & 42.83
& 42.0 & 46.0 & 33.0 & 77.0 & 30.0 & 47.0 & 45.83
& 44.33 \\

4DThinker-3B~\citep{chen20264dthinker}
& 43.0 & 42.0 & 61.0 & 52.0 & 32.0 & 34.0 & 44.00
& 50.0 & 37.0 & 56.0 & 41.0 & 29.0 & 26.0 & 39.83
& 41.92 \\

\midrule

\textbf{Trace4D (Qwen3-VL-4B)}
& \textbf{79.0} & \textbf{71.0} & 85.0 & 61.0 & \textbf{97.0} & \textbf{93.0} & 81.00
& \textbf{73.0} & 59.0 & 75.0 & 95.0 & \textbf{93.0} & \textbf{70.0} & \textbf{77.50}
& \textbf{79.25} \\

\textbf{Trace4D (Qwen3-VL-8B)}
& \textbf{79.0} & 69.0 & \textbf{89.0} & \textbf{68.0} & 96.0 & 86.0 & \textbf{81.17}
& 68.0 & \textbf{63.0} & \textbf{77.0} & \textbf{96.0} & 92.0 & 65.0 & 76.83
& 79.00 \\

\bottomrule
\end{tabular}
}
\caption{Comparison of vision-language models on Trace4D-Bench, covering relative and absolute spatial and motion understanding.}
\label{tab:main_results}
\end{table*}

\subsection{Results on Trace4D-Bench}
Trace4D-Bench evaluates distance, direction, speed, and translation under specified relative and absolute coordinate frames. Its questions use the same task formulation as Trace4D-Train, with source videos held out from training.

We compare three baseline groups: proprietary models GPT-4o~\citep{hurst2024gpt} and GPT-5~\citep{singh2025openai}; general-purpose VLMs Qwen2.5-VL~\citep{Bai2025Qwen25VLTR}, Qwen3-VL~\citep{bai2025qwen3}, InternVL2.5/3.5~\citep{chen2024expanding,wang2025internvl3}, and LLaVA-OneVision1.5~\citep{an2025llava}; and 3D/4D-specialized models VG-LLM~\citep{zheng2026learning}, VLM-3R~\citep{fan2026vlm}, MLLM-4D~\citep{yin2026mllm}, GSM~\citep{zhou2026learning}, and 4DThinker~\citep{chen20264dthinker}. Table~\ref{tab:main_results} reports task-wise QA accuracy (\%) across the spatial and motion reasoning categories.

Trace4D achieves overall accuracies of $79.25\%$ and $79.00\%$ with the 4B and 8B models, respectively, outperforming the evaluated proprietary, general-purpose, and 3D/4D-specialized baselines. Both models improve over their corresponding base VLMs by more than $37$ percentage points. Each model also outperforms the evaluated baselines across all 12 question types, covering both relative and absolute reasoning on unseen videos. The ablations in Section~\ref{sec:ablation_analysis} further examine the roles of reasoning over 4D state changes and incorporating temporal visual observations.

\subsection{Results on VLM4D}
\begin{wraptable}{R}{0.46\textwidth}
\vspace{-14pt}
\centering
\fontsize{7}{8}\selectfont
\setlength{\tabcolsep}{2pt}
\begin{tabular*}{\linewidth}{@{\extracolsep{\fill}}lccc@{}}
\toprule
\textbf{Model} & \textbf{Syn.} & \textbf{Real.} & \textbf{Avg.} \\
\midrule
\multicolumn{4}{l}{\textit{Proprietary Models}} \\
GPT-4o~\citep{hurst2024gpt} & 54.38 & 56.02 & 55.62 \\
GPT-5~\citep{singh2025openai} & 62.70 & 74.69 & 71.75 \\
\midrule
\multicolumn{4}{l}{\textit{General VLMs}} \\
Qwen2.5-VL-3B~\citep{Bai2025Qwen25VLTR} & 35.96 & 47.34 & 44.55 \\
Qwen3-VL-4B~\citep{bai2025qwen3} & 48.76 & 57.55 & 55.40 \\
Qwen2.5-VL-7B & 48.31 & 51.79 & 50.94 \\
Qwen3-VL-8B & 55.73 & \underline{60.25} & 59.14 \\
InternVL2.5-4B~\citep{chen2024expanding} & 32.81 & 50.84 & 46.42 \\
InternVL3.5-4B~\citep{wang2025internvl3} & 42.24 & 56.38 & 52.92 \\
InternVL2.5-8B & 40.45 & 55.80 & 52.04 \\
InternVL3.5-8B & 46.07 & 55.87 & 53.47 \\
LLaVA-OneVision1.5-4B~\citep{an2025llava} & 40.00 & 44.42 & 43.34 \\
LLaVA-OneVision1.5-8B & 40.45 & 45.22 & 44.05 \\
\midrule
\multicolumn{4}{l}{\textit{3D/4D VLMs}} \\
VG-LLM-8B~\citep{zheng2026learning} & 34.83 & 48.87 & 45.43 \\
VLM-3R-7B~\citep{fan2026vlm} & 35.28 & 52.88 & 48.57 \\
MLLM-4D-8B~\citep{yin2026mllm} & 55.05 & \textbf{63.09} & 61.12 \\
GSM-7B~\citep{zhou2026learning} & 50.11 & 47.56 & 48.18 \\
4DThinker-3B~\citep{chen20264dthinker} & 24.72 & 38.95 & 35.46 \\
\textbf{Trace4D (Qwen3-VL-4B)} & \textbf{77.30} & 60.10 & \textbf{64.32} \\
\textbf{Trace4D (Qwen3-VL-8B)} & \underline{68.09} & 59.08 & \underline{61.29} \\
\bottomrule
\end{tabular*}
\caption{Performance on VLM4D. \textbf{Bold} and \underline{underline} denote the best and second-best scores among non-proprietary models.}
\label{tab:vlm4d}
\vspace{-10pt}
\end{wraptable}

VLM4D~\citep{zhou2025vlm4d} evaluates spatiotemporal reasoning in real and synthetic videos, including first- and third-person views. Its multiple-choice questions target motion, perspective, and temporal continuity in the videos, while Trace4D-Bench constructs questions from tracked 4D states under specified coordinate conventions. Their distinct video collections and question formulations provide complementary evaluations of 4D understanding.

As shown in Table~\ref{tab:vlm4d}, Trace4D reaches $64.32\%$ with the 4B model and $61.29\%$ with the 8B model, exceeding their respective base VLMs by $8.92$ and $2.15$ percentage points. Both models outperform the evaluated general-purpose VLMs in overall accuracy. The gains over the base VLMs are larger on synthetic videos, while real-video accuracy improves for 4B but decreases for 8B. These results indicate transfer to a different video collection and question formulation, with gains that vary across model sizes and video subsets. Notably, the 4B model outperforms the 8B model on VLM4D and achieves a substantially larger improvement over its base VLM. One possible explanation is a capacity–data mismatch: the fixed amount of task-specific supervision may not fully exploit the larger model’s capacity, with this limitation becoming more apparent under the distribution shift in VLM4D.

\subsection{Ablation and Analysis}
\label{sec:ablation_analysis}
\paragraph{CoT ablation.}

\begin{wraptable}{r}{0.49\textwidth}
\vspace{-\intextsep}
\centering
\scriptsize
\setlength{\tabcolsep}{3pt}
\renewcommand{\arraystretch}{1.0}
\begin{tabular*}{\linewidth}{@{\extracolsep{\fill}}lcc@{}}
\toprule
\textbf{CoT Variant} & \textbf{Trace4D-Bench} & \textbf{VLM4D} \\
\midrule
State-only & 74.83 & 52.86 \\
4D-state CoT & \textbf{79.42} & 54.68 \\
Visual-only CoT & 73.75 & \underline{59.25} \\
\textbf{4D + Visual CoT} & \underline{77.25} & \textbf{60.52} \\
\bottomrule
\end{tabular*}
\caption{CoT ablation. Bold and underlined values indicate the best and second-best results, respectively.}
\label{tab:cot_ablation}
\vspace{-10pt}
\end{wraptable}

We compare four response formulations to examine the roles of reasoning over 4D state changes and incorporating temporal visual observations. All variants start from the same perception-pretrained 4B checkpoint and undergo matched QA adaptation without GRPO. Only the QA response supervision and task instructions differ; QA pairs, perception replay, and optimization settings are fixed. This controlled setting is distinct from the final method's joint SFT. \textit{State-only} predicts the structured state sequence within \texttt{<4d>} and directly answers. \textit{4D-state CoT} adds reasoning over the predicted state changes, without explicit visual analysis. \textit{Visual-only CoT} reasons from video observations without generating a \texttt{<4d>} block. Our \textit{4D + Visual CoT} connects predicted 4D state changes with temporal visual observations of the queried entities over matching time intervals. Table~\ref{tab:cot_ablation} reports both benchmarks; Appendix~\ref{app:cot_ablation} describes the shared SFT schedule and perception replay.

Supervising reasoning over 4D state changes improves QA accuracy beyond state prediction alone. Compared with \textit{State-only}, \textit{4D-state CoT} improves accuracy by $4.59$ percentage points on Trace4D-Bench and $1.82$ points on VLM4D. This comparison supports supervising how state changes lead to an answer, in addition to supervising the states themselves.

Combining state-based reasoning with temporal visual analysis yields the highest VLM4D accuracy among the four variants. Our \textit{4D + Visual CoT} achieves $77.25\%$ on Trace4D-Bench and $60.52\%$ on VLM4D, improving over \textit{Visual-only CoT} on both benchmarks. While \textit{4D-state CoT} performs best on Trace4D-Bench ($79.42\%$), adding temporal visual analysis raises VLM4D accuracy from $54.68\%$ to $60.52\%$. These results suggest that temporal visual observations provide complementary evidence for interpreting 4D state changes.

\begin{figure}[t]
    \centering
    \begin{minipage}[t]{0.44\textwidth}
        \vspace{0pt}
        \centering
        \includegraphics[width=0.893\linewidth]{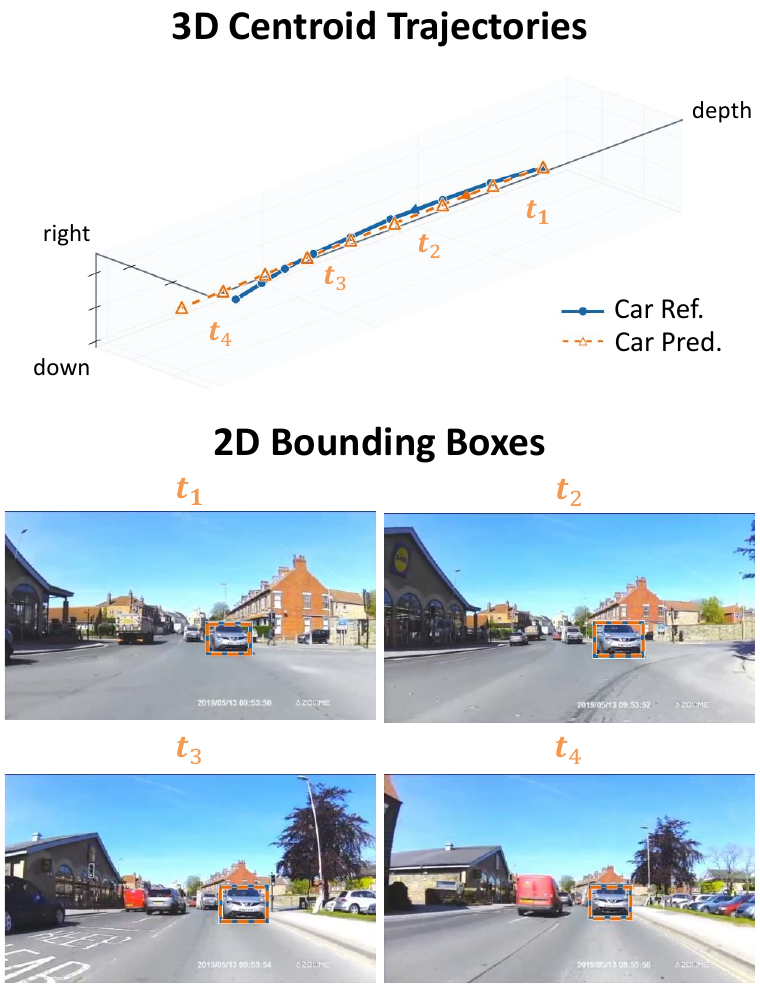}
        \caption{Predicted 3D centroid trajectories and 2D bounding boxes across timesteps.}
        \label{fig:4d_visualization}
    \end{minipage}
    \hfill
    \begin{minipage}[t]{0.52\textwidth}
        \vspace{0pt}
        \centering
        \begin{minipage}[t]{\linewidth}
\vspace{0pt}
\centering
\footnotesize
\setlength{\tabcolsep}{3pt}
\renewcommand{\arraystretch}{1.0}
\begin{tabular*}{\linewidth}{@{\extracolsep{\fill}}lccc@{}}
\toprule
\textbf{Training Strategy} & \textbf{Rel.} & \textbf{Abs.} & \textbf{VLM4D} \\
\midrule
Adaptation-only SFT & 76.33 & 67.50 & 57.54 \\
Perception-first SFT & 79.17 & 75.33 & 60.52 \\
Joint SFT & 79.33 & 76.33 & 62.39 \\
\textbf{Joint SFT + GRPO (Trace4D)} & \textbf{81.00} & \textbf{77.50} & \textbf{64.32} \\
\bottomrule
\end{tabular*}
\captionof{table}{Training strategy comparison.}
\label{tab:stagewise_ablation}
\end{minipage}

        \par\vspace{9pt}
        \begin{minipage}[t]{\linewidth}
\vspace{0pt}
\centering
\footnotesize
\setlength{\tabcolsep}{3pt}
\renewcommand{\arraystretch}{1.0}
\begin{tabular*}{\linewidth}{@{\extracolsep{\fill}}lccc@{}}
\toprule
\textbf{Setting} & \textbf{IoU} $\uparrow$ & \textbf{Centroid Error} $\downarrow$ & \textbf{Trajectory L2} $\downarrow$ \\
\midrule
Relative & 0.7556 & 0.0994 & 0.0852 \\
Absolute & 0.7176 & 0.2654 & 0.1428 \\
\bottomrule
\end{tabular*}
\captionof{table}{Trace4D (4B) state prediction.}
\label{tab:perception_eval}
\end{minipage}

        \par\vspace{9pt}
        \begin{minipage}{\linewidth}
\centering
\footnotesize
\setlength{\tabcolsep}{3pt}
\renewcommand{\arraystretch}{1.0}
\begin{tabular*}{\linewidth}{@{\extracolsep{\fill}}lccc@{}}
\toprule
\textbf{Setting} & \textbf{Low L2} & \textbf{Mid L2} & \textbf{High L2} \\
\midrule
Relative & 89.90 & 81.37 & 71.72 \\
Absolute & 91.92 & 75.00 & 65.66 \\
\bottomrule
\end{tabular*}
\captionof{table}{QA accuracy (\%) by L2 error group.}
\label{tab:perception_qa}
\end{minipage}

    \end{minipage}
    \vspace{-10pt}
\end{figure}

\paragraph{Training strategy.}
Table~\ref{tab:stagewise_ablation} compares three SFT recipes and the GRPO continuation of Joint SFT. \textit{Adaptation-only SFT} performs QA adaptation from the base model without perception supervision. \textit{Perception-first SFT} first performs one epoch of perception pretraining, followed by QA adaptation with perception replay. \textit{Joint SFT} jointly mixes perception and grounded QA supervision from the base model. Details are provided in Appendix~\ref{app:stagewise_ablation}.

Adding perception supervision substantially improves over \textit{Adaptation-only SFT}. \textit{Joint SFT} further improves over \textit{Perception-first SFT} across all three evaluation settings, indicating a benefit from jointly learning perception and grounded QA. Continuing Joint SFT with GRPO on 7,008 QA examples further improves performance in both coordinate settings and on VLM4D.

\paragraph{Perception and reasoning analysis.}
We evaluate 4D perception on clips held out at the source-video level. Table~\ref{tab:perception_eval} reports bounding-box IoU, centroid error, and trajectory L2, defined in Appendix~\ref{app:evaluation_protocol}. Absolute evaluation, in which object states must be expressed relative to a fixed anchor viewpoint, yields larger centroid and trajectory errors than Relative evaluation. Figure~\ref{fig:4d_visualization} shows a predicted car trajectory following the reference with deviations at individual timesteps.

We examine the association between trajectory L2 error and QA accuracy. Within each question type, we divide responses into three approximately equal-sized groups by error and then average QA accuracy across question types. Table~\ref{tab:perception_qa} shows that QA accuracy decreases as trajectory error increases in both settings, with a larger low-to-high error gap in the Absolute setting ($26.26$ vs. $18.18$ percentage points). This pattern suggests a connection between 4D perception quality and downstream reasoning performance.
\section{Conclusion}

We presented Trace4D, a framework for grounded spatiotemporal reasoning via predicted 4D state sequences. Trace4D predicts temporally ordered 4D states and uses their inter-timestep changes together with temporal visual observations as intermediate evidence for reasoning. Experiments on Trace4D-Bench and VLM4D show improved spatiotemporal reasoning, while ablations demonstrate the benefits of explicit 4D state-change reasoning, complementary temporal visual evidence, and geometry-aware optimization. Our analysis further reveals a connection between 4D perception quality and downstream QA performance, highlighting the role of explicit 4D states in bridging perception and reasoning.

\subsection*{AI use statement}

In this work, we used generative AI tools to generate chain-of-thought (CoT) annotations for our training data. We have not used generative AI tools for other research tasks requiring disclosure, including proposing or refining hypotheses or interpreting results. Additionally, we used generative AI tools to improve the grammar, readability, and flow of the manuscript. All AI-assisted work was reviewed and verified by the authors, and we take full responsibility for the final content of this work, including all text, claims, code, and other artifacts produced with the aid of generative AI.

\section*{Reproducibility Statement}
We describe dataset construction, coordinate normalization, and chain-of-thought generation in Appendix~\ref{app:dataset_construction}. Training hyperparameters, geometry-aware GRPO rewards, and evaluation protocols are detailed in Appendix~\ref{app:training_details}, including the model-specific reward weights and decoding settings. Training and inference prompts are provided in Appendix~\ref{app:shared_prompts}. We plan to publicly release our code and trained model checkpoints upon publication to facilitate reproduction of our results.

\bibliography{iclr2027_conference}

@inproceedings{yang2025thinking,
  title={Thinking in space: How multimodal large language models see, remember, and recall spaces},
  author={Yang, Jihan and Yang, Shusheng and Gupta, Anjali W and Han, Rilyn and Fei-Fei, Li and Xie, Saining},
  booktitle={2025 IEEE/CVF Conference on Computer Vision and Pattern Recognition (CVPR)},
  pages={10632--10643},
  year={2025},
  organization={IEEE}
}

@inproceedings{zhou2026learning,
  title={Learning to reason in 4d: Dynamic spatial understanding for vision language models},
  author={Zhou, Shengchao and Chen, Yuxin and Ge, Yuying and Huang, Wei and Lin, Jiehong and Shan, Ying and Qi, Xiaojuan},
  booktitle={Proceedings of the IEEE/CVF Conference on Computer Vision and Pattern Recognition},
  pages={9637--9646},
  year={2026}
}

@inproceedings{zhou2025vlm4d,
  title={Vlm4d: Towards spatiotemporal awareness in vision language models},
  author={Zhou, Shijie and Vilesov, Alexander and He, Xuehai and Wan, Ziyu and Zhang, Shuwang and Nagachandra, Aditya and Chang, Di and Chen, Dongdong and Wang, Xin Eric and Kadambi, Achuta},
  booktitle={2025 IEEE/CVF International Conference on Computer Vision (ICCV)},
  pages={8600--8612},
  year={2025},
  organization={IEEE}
}

@inproceedings{fan2026vlm,
  title={Vlm-3r: Vision-language models augmented with instruction-aligned 3d reconstruction},
  author={Fan, Zhiwen and Zhang, Jian and Li, Renjie and Zhang, Junge and Chen, Runjin and Hu, Hezhen and Wang, Kevin and Wang, Peihao and Qu, Huaizhi and Zhou, Shijie and others},
  booktitle={Proceedings of the IEEE/CVF Conference on Computer Vision and Pattern Recognition},
  pages={31054--31065},
  year={2026}
}

@article{an2025llava,
  title={Llava-onevision-1.5: Fully open framework for democratized multimodal training},
  author={An, Xiang and Xie, Yin and Yang, Kaicheng and Zhang, Wenkang and Zhao, Xiuwei and Cheng, Zheng and Wang, Yirui and Xu, Songcen and Chen, Changrui and Zhu, Didi and others},
  journal={arXiv preprint arXiv:2509.23661},
  year={2025}
}

@article{zheng2026learning,
  title={Learning from videos for 3d world: Enhancing mllms with 3d vision geometry priors},
  author={Zheng, Duo and Li, Yanyang and Wang, Liwei and others},
  journal={Advances in neural information processing systems},
  volume={38},
  pages={20560--20586},
  year={2026}
}

@article{chen2024expanding,
  title={Expanding performance boundaries of open-source multimodal models with model, data, and test-time scaling},
  author={Chen, Zhe and Wang, Weiyun and Cao, Yue and Liu, Yangzhou and Gao, Zhangwei and Cui, Erfei and Zhu, Jinguo and Ye, Shenglong and Tian, Hao and Liu, Zhaoyang and others},
  journal={arXiv preprint arXiv:2412.05271},
  year={2024}
}

@article{liu2026geoalign,
  title={GeoAlign: Geometric Feature Realignment for MLLM Spatial Reasoning},
  author={Liu, Zhaochen and Qiao, Limeng and Wan, Guanglu and Jiang, Tingting},
  journal={arXiv preprint arXiv:2604.12630},
  year={2026}
}

@article{yin2026mllm,
  title={Mllm-4d: Towards visual-based spatial-temporal intelligence},
  author={Yin, Xingyilang and Li, Chengzhengxu and Chang, Jiahao and Pun, Chi-Man and Cun, Xiaodong},
  journal={arXiv preprint arXiv:2603.00515},
  year={2026}
}

@article{cho20264dp,
  title={4dp-qa: Scalable qa for 4d perception in vision language models},
  author={Cho, Seokju and Badki, Abhishek and Su, Hang and Jiang, Jindong and Zeng, Ziyao and Kim, Seungryong and Liu, Sifei and Gallo, Orazio},
  journal={arXiv preprint arXiv:2606.11568},
  year={2026}
}

@article{chen20264dthinker,
  title={4DThinker: Thinking with 4D Imagery for Dynamic Spatial Understanding},
  author={Chen, Zhangquan and Zhang, Manyuan and Yu, Xinlei and An, Xiang and Li, Bo and Xie, Xin and Wang, ZiDong and Sun, Mingze and Chen, Shuang and Li, Hongyu and others},
  journal={arXiv preprint arXiv:2605.05997},
  year={2026}
}

@article{wang2025video,
  title={Video-str: Reinforcing mllms in video spatio-temporal reasoning with relation graph},
  author={Wang, Wentao and Zou, Heqing and Luo, Tianze and Huang, Rui and Zhao, Yutian and Wang, Zhuochen and Zhang, Hansheng and Qin, Chengwei and Wang, Yan and Zhao, Lin and others},
  journal={arXiv preprint arXiv:2510.10976},
  year={2025}
}

@inproceedings{wu2026grounded,
  title={Grounded chain-of-thought for multimodal large language models},
  author={Wu, Qiong and Yang, Xiangcong and Zhou, Yiyi and Fang, Chenxin and Song, Baiyang and Sun, Xiaoshuai and Ji, Rongrong},
  booktitle={Proceedings of the IEEE/CVF Conference on Computer Vision and Pattern Recognition},
  pages={33577--33587},
  year={2026}
}

@inproceedings{wang2026traceable,
  title={Traceable evidence enhanced visual grounded reasoning: Evaluation and method},
  author={Wang, Haochen and Li, Xiangtai and Huang, Zilong and Wang, Anran and Wang, Jiacong and Zhang, Tao and Bai, Sule and Kang, Zijian and Feng, Jiashi and Wang, Zhuochen and others},
  booktitle={International Conference on Learning Representations},
  volume={2026},
  pages={148769--148794},
  year={2026}
}

@article{meng2025open,
  title={Open-o3-Video: Grounded Video Reasoning with Explicit Spatio-Temporal Evidence},
  author={Meng, Jiahao and Li, Xiangtai and Wang, Haochen and Tan, Yue and Zhang, Tao and Kong, Lingdong and Tong, Yunhai and Wang, Anran and Teng, Zhiyang and Wang, Yujing and others},
  journal={arXiv preprint arXiv:2510.20579},
  year={2025}
}

@inproceedings{wang2026pi,
  title={$\pi^3$: Permutation-Equivariant Visual Geometry Learning},
  author={Wang, Yifan and Zhou, Jianjun and Zhu, Haoyi and Chang, Wenzheng and Zhou, Yang and Li, Zizun and Chen, Junyi and Pang, Jiangmiao and Shen, Chunhua and He, Tong},
  booktitle={International Conference on Learning Representations},
  volume={2026},
  pages={10481--10497},
  year={2026}
}

@article{ren2024grounded,
  title={Grounded sam: Assembling open-world models for diverse visual tasks},
  author={Ren, Tianhe and Liu, Shilong and Zeng, Ailing and Lin, Jing and Li, Kunchang and Cao, He and Chen, Jiayu and Huang, Xinyu and Chen, Yukang and Yan, Feng and others},
  journal={arXiv preprint arXiv:2401.14159},
  year={2024}
}

@inproceedings{wang2025koala,
  title={Koala-36m: A large-scale video dataset improving consistency between fine-grained conditions and video content},
  author={Wang, Qiuheng and Shi, Yukai and Ou, Jiarong and Chen, Rui and Lin, Ke and Wang, Jiahao and Jiang, Boyuan and Yang, Haotian and Zheng, Mingwu and Tao, Xin and others},
  booktitle={2025 IEEE/CVF Conference on Computer Vision and Pattern Recognition (CVPR)},
  pages={8428--8437},
  year={2025},
  organization={IEEE}
}

@article{bai2025qwen3,
  title={Qwen3-vl technical report},
  author={Bai, Shuai and Cai, Yuxuan and Chen, Ruizhe and Chen, Keqin and Chen, Xionghui and Cheng, Zesen and Deng, Lianghao and Ding, Wei and Gao, Chang and Ge, Chunjiang and others},
  journal={arXiv preprint arXiv:2511.21631},
  year={2025}
}

@article{wang2025internvl3,
  title={Internvl3.5: Advancing open-source multimodal models in versatility, reasoning, and efficiency},
  author={Wang, Weiyun and Gao, Zhangwei and Gu, Lixin and Pu, Hengjun and Cui, Long and Wei, Xingguang and Liu, Zhaoyang and Jing, Linglin and Ye, Shenglong and Shao, Jie and others},
  journal={arXiv preprint arXiv:2508.18265},
  year={2025}
}

@article{hurst2024gpt,
  title={{GPT-4o System Card}},
  author={Hurst, Aaron and Lerer, Adam and Goucher, Adam P and Perelman, Adam and Ramesh, Aditya and Clark, Aidan and Ostrow, AJ and Welihinda, Akila and Hayes, Alan and Radford, Alec and others},
  journal={arXiv preprint arXiv:2410.21276},
  year={2024}
}

@article{singh2025openai,
  title={{OpenAI GPT-5 System Card}},
  author={Singh, Aaditya and Fry, Adam and Perelman, Adam and Tart, Adam and Ganesh, Adi and El-Kishky, Ahmed and McLaughlin, Aidan and Low, Aiden and Ostrow, AJ and Ananthram, Akhila and others},
  journal={arXiv preprint arXiv:2601.03267},
  year={2025}
}

@article{gao2026dyntrace,
  title={DynTrace: Tracking Dynamic Object Evidence for 4D Spatio-Temporal Reasoning in MLLMs},
  author={Gao, Rongxin and Huang, Yuzhi and Liu, Dongxuan and Li, Chu and Wang, Zhenye and Wu, Jie and Xie, Shuzhao and Jiang, Jingyan and Ding, Xinghao and Tu, Xiaotong and others},
  journal={arXiv preprint arXiv:2607.12503},
  year={2026}
}

@article{zhang2024llava,
  title={Llava-video: Video instruction tuning with synthetic data},
  author={Zhang, Yuanhan and Wu, Jinming and Li, Wei and Li, Bo and Ma, Zejun and Liu, Ziwei and Li, Chunyuan},
  journal={arXiv preprint arXiv:2410.02713},
  year={2024}
}

@article{Bai2025Qwen25VLTR,
  title={Qwen2.5-VL Technical Report},
  author={Bai, Shuai and Chen, Keqin and Liu, Xuejing and Wang, Jialin and Ge, Wenbin and Song, Sibo and Dang, Kai and Wang, Peng and Wang, Shijie and Tang, Jun and Zhong, Humen and Zhu, Yuanzhi and Yang, Mingkun and Li, Zhaohai and Wan, Jianqiang and Wang, Pengfei and Ding, Wei and Fu, Zheren and Xu, Yiheng and Ye, Jiabo and Zhang, Xi and Xie, Tianbao and Cheng, Zesen and Zhang, Hang and Yang, Zhibo and Xu, Haiyang and Lin, Junyang},
  journal={arXiv preprint arXiv:2502.13923},
  year={2025}
}

@inproceedings{wang2026vgr,
  title={Vgr: Visual grounded reasoning},
  author={Wang, Jiacong and Kang, Zijian and Wang, Haochen and Liang, Xiao and Wang, Ya and Li, Jiawen and Wu, Bohong and Ran, Jiao and Jiang, Haiyong and Feng, Chao and others},
  booktitle={International Conference on Learning Representations},
  volume={2026},
  pages={89506--89528},
  year={2026}
}

@article{sarch2026grounded,
  title={Grounded reinforcement learning for visual reasoning},
  author={Sarch, Gabriel and Saha, Snigdha and Khandelwal, Naitik and Jain, Ayush and Tarr, Michael and Kumar, Aviral and Fragkiadaki, Katerina},
  journal={Advances in Neural Information Processing Systems},
  volume={38},
  pages={150977--151013},
  year={2026}
}

@article{yang2026cambrian,
  title={Cambrian-P: Pose-grounded video understanding},
  author={Yang, Jihan and Zhao, Zifan and Pan, Xichen and Yang, Shusheng and Zhang, Junyi and Kang, Bingyi and Xu, Hu and Li, Shang-Wen and Xie, Saining},
  journal={arXiv preprint arXiv:2605.22819},
  year={2026}
}

@inproceedings{chen2026reasoning,
  title={Reasoning in space via grounding in the world},
  author={Chen, Yiming and Qi, Zekun and Zhang, Wenyao and Jin, Xin and Zhang, Li and Liu, Peidong},
  booktitle={International Conference on Learning Representations},
  volume={2026},
  pages={65549--65570},
  year={2026}
}

@inproceedings{linghu2026scenecot,
  title={Scenecot: Eliciting grounded chain-of-thought reasoning in 3d scenes},
  author={Linghu, Xiongkun and Huang, Jiangyong and Zhu, Ziyu and Jia, Baoxiong and Huang, Siyuan},
  booktitle={International Conference on Learning Representations},
  volume={2026},
  pages={2754--2779},
  year={2026}
}

@inproceedings{deng2025motion,
  title={Motion-grounded video reasoning: Understanding and perceiving motion at pixel level},
  author={Deng, Andong and Chen, Tongjia and Yu, Shoubin and Yang, Taojiannan and Spencer, Lincoln and Tian, Yapeng and Mian, Ajmal Saeed and Bansal, Mohit and Chen, Chen},
  booktitle={2025 IEEE/CVF Conference on Computer Vision and Pattern Recognition (CVPR)},
  pages={8625--8636},
  year={2025},
  organization={IEEE}
}

@article{cheng2024spatialrgpt,
  title={Spatialrgpt: Grounded spatial reasoning in vision-language models},
  author={Cheng, An-Chieh and Yin, Hongxu and Fu, Yang and Guo, Qiushan and Yang, Ruihan and Kautz, Jan and Wang, Xiaolong and Liu, Sifei},
  journal={Advances in Neural Information Processing Systems},
  volume={37},
  pages={135062--135093},
  year={2024}
}

@article{miao2026geoweaver,
  title={GeoWeaver: Grounding Visual Tokens with Geometric Evidence before Scene Reasoning},
  author={Miao, Deshui and Huang, Xingsen and Gu, Yameng and Li, Xin and Zhang, Haijun and Yang, Ming-Hsuan},
  journal={arXiv preprint arXiv:2605.22558},
  year={2026}
}

@article{shao2024deepseekmath,
  title={Deepseekmath: Pushing the limits of mathematical reasoning in open language models},
  author={Shao, Zhihong and Wang, Peiyi and Zhu, Qihao and Xu, Runxin and Song, Junxiao and Bi, Xiao and Zhang, Haowei and Zhang, Mingchuan and Li, YK and Wu, Yang and others},
  journal={arXiv preprint arXiv:2402.03300},
  year={2024}
}

@article{wei2022chain,
  title={Chain-of-thought prompting elicits reasoning in large language models},
  author={Wei, Jason and Wang, Xuezhi and Schuurmans, Dale and Bosma, Maarten and Xia, Fei and Chi, Ed and Le, Quoc V and Zhou, Denny and others},
  journal={Advances in neural information processing systems},
  volume={35},
  pages={24824--24837},
  year={2022}
}

@article{lu2022learn,
  title={Learn to explain: Multimodal reasoning via thought chains for science question answering},
  author={Lu, Pan and Mishra, Swaroop and Xia, Tanglin and Qiu, Liang and Chang, Kai-Wei and Zhu, Song-Chun and Tafjord, Oyvind and Clark, Peter and Kalyan, Ashwin},
  journal={Advances in neural information processing systems},
  volume={35},
  pages={2507--2521},
  year={2022}
}

@article{zhang2023multimodal,
  title={Multimodal chain-of-thought reasoning in language models},
  author={Zhang, Zhuosheng and Zhang, Aston and Li, Mu and Zhao, Hai and Karypis, George and Smola, Alex},
  journal={arXiv preprint arXiv:2302.00923},
  year={2023}
}

@inproceedings{xu2025llava,
  title={Llava-cot: Let vision language models reason step-by-step},
  author={Xu, Guowei and Jin, Peng and Wu, Ziang and Li, Hao and Song, Yibing and Sun, Lichao and Yuan, Li},
  booktitle={2025 IEEE/CVF International Conference on Computer Vision (ICCV)},
  pages={2087--2098},
  year={2025},
  organization={IEEE}
}

@article{shao2024visual,
  title={Visual cot: Advancing multi-modal language models with a comprehensive dataset and benchmark for chain-of-thought reasoning},
  author={Shao, Hao and Qian, Shengju and Xiao, Han and Song, Guanglu and Zong, Zhuofan and Wang, Letian and Liu, Yu and Li, Hongsheng},
  journal={Advances in Neural Information Processing Systems},
  volume={37},
  pages={8612--8642},
  year={2024}
}

@inproceedings{li2025vocot,
  title={Vocot: Unleashing visually grounded multi-step reasoning in large multi-modal models},
  author={Li, Zejun and Luo, Ruipu and Zhang, Jiwen and Qiu, Minghui and Huang, Xuan-Jing and Wei, Zhongyu},
  booktitle={Proceedings of the 2025 Conference of the Nations of the Americas Chapter of the Association for Computational Linguistics: Human Language Technologies (Volume 1: Long Papers)},
  pages={3769--3798},
  year={2025}
}

@inproceedings{han2025videoespresso,
  title={Videoespresso: A large-scale chain-of-thought dataset for fine-grained video reasoning via core frame selection},
  author={Han, Songhao and Huang, Wei and Shi, Hairong and Zhuo, Le and Su, Xiu and Zhang, Shifeng and Zhou, Xu and Qi, Xiaojuan and Liao, Yue and Liu, Si},
  booktitle={2025 IEEE/CVF Conference on Computer Vision and Pattern Recognition (CVPR)},
  pages={26181--26191},
  year={2025},
  organization={IEEE}
}

@article{hu2021lora,
  title={Lora: Low-rank adaptation of large language models},
  author={Hu, Edward J and Shen, Yelong and Wallis, Phillip and Allen-Zhu, Zeyuan and Li, Yuanzhi and Wang, Shean and Wang, Lu and Chen, Weizhu},
  journal={arXiv preprint arXiv:2106.09685},
  year={2021}
}

@inproceedings{chen2024spatialvlm,
  title={Spatialvlm: Endowing vision-language models with spatial reasoning capabilities},
  author={Chen, Boyuan and Xu, Zhuo and Kirmani, Sean and Ichter, Brian and Sadigh, Dorsa and Guibas, Leonidas and Xia, Fei},
  booktitle={Proceedings of the IEEE/CVF conference on computer vision and pattern recognition},
  pages={14455--14465},
  year={2024}
}

@article{ko2025st,
  title={St-vlm: Kinematic instruction tuning for spatio-temporal reasoning in vision-language models},
  author={Ko, Dohwan and Kim, Sihyeon and Suh, Yumin and Yoon, Minseo and Chandraker, Manmohan and Kim, Hyunwoo J and others},
  journal={arXiv preprint arXiv:2503.19355},
  year={2025}
}

@inproceedings{xu2026multi,
  title={Multi-spatialmllm: Multi-frame spatial understanding with multi-modal large language models},
  author={Xu, Runsen and Wang, Weiyao and Tang, Hao and Chen, Xingyu and Wang, Xiaodong and Chu, Fu-Jen and Feiszli, Matt and Liang, Kevin J},
  booktitle={Proceedings of the IEEE/CVF Conference on Computer Vision and Pattern Recognition},
  pages={31078--31088},
  year={2026}
}

@article{ma2026spatialreasoner,
  title={Spatialreasoner: Towards explicit and generalizable 3d spatial reasoning},
  author={Ma, Wufei and Chou, Yu-Cheng and Liu, Qihao and Wang, Xingrui and de Melo, Celso and Xie, Jianwen and Yuille, Alan},
  journal={Advances in Neural Information Processing Systems},
  volume={38},
  pages={140751--140774},
  year={2026}
}

@article{ouyang2025spacer,
  title={Spacer: Reinforcing mllms in video spatial reasoning},
  author={Ouyang, Kun and Liu, Yuanxin and Wu, Haoning and Liu, Yi and Zhou, Hao and Zhou, Jie and Meng, Fandong and Sun, Xu},
  journal={arXiv preprint arXiv:2504.01805},
  year={2025}
}

@inproceedings{ravi2025sam,
  title={Sam 2: Segment anything in images and videos},
  author={Ravi, Nikhila and Gabeur, Valentin and Hu, Yuan-Ting and Hu, Ronghang and Ryali, Chaitanya and Ma, Tengyu and Khedr, Haitham and R{\"a}dle, Roman and Rolland, Chloe and Gustafson, Laura and others},
  booktitle={International Conference on Learning Representations},
  volume={2025},
  pages={28085--28128},
  year={2025}
}

@inproceedings{liu2024grounding,
  title={Grounding dino: Marrying dino with grounded pre-training for open-set object detection},
  author={Liu, Shilong and Zeng, Zhaoyang and Ren, Tianhe and Li, Feng and Zhang, Hao and Yang, Jie and Jiang, Qing and Li, Chunyuan and Yang, Jianwei and Su, Hang and others},
  booktitle={European conference on computer vision},
  pages={38--55},
  year={2024},
  organization={Springer}
}
\bibliographystyle{iclr2027_conference}

\clearpage
\appendix
\section*{\LARGE Appendix}
\section{Data Generation and Dataset Details}
\label{app:dataset_construction}

\subsection{State Representation and Coordinate Normalization}
\label{app:state_serialization}
\label{app:coordinate_normalization}

Each video input contains 32 sampled frames, indexed from 0 to 31. Reference states may cover a subset of these timesteps; state and anchor frame indices use the same sampled timeline. The reference 4D state sequence is serialized as a time-ordered JSON array within \texttt{<4d>}, with \texttt{frame} and \texttt{entities} fields. Object entries contain \texttt{label}, \texttt{bbox\_2d}, and \texttt{centroid}. Camera entries, when required for absolute tasks, use the label \texttt{cam} and omit the \texttt{bbox\_2d} field.

Bounding-box coordinates are normalized by image width and height and scaled to $[0,1000]$. The 3D axes point right, down, and forward. For perception examples and spatial QA, 3D positions are divided by the largest positive depth of the referenced objects at the first clip frame. Motion QA uses the median valid positive object depth within the selected window. Each example uses one scale across all targets and annotated timesteps, yielding dimensionless 3D coordinates.

\subsection{QA Construction}
\label{app:qa_templates}

\begin{wrapfigure}[25]{r}{0.47\linewidth}
    \vspace{-25pt}
    \centering
    \includegraphics[width=\linewidth]{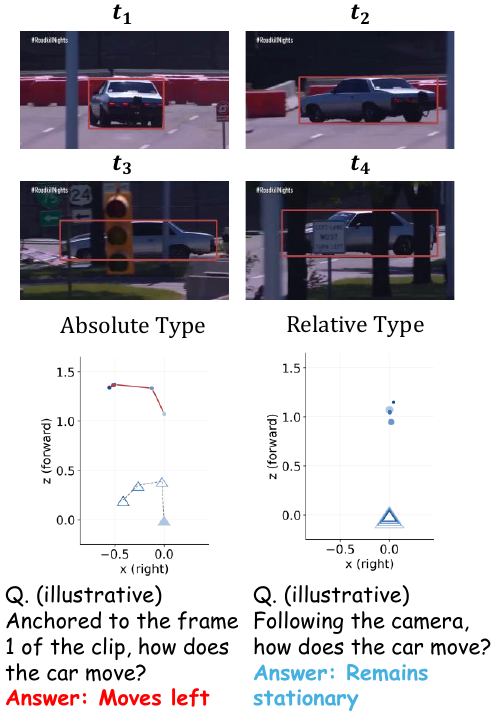}
    \caption{Relative vs.\ Absolute frames for the same clip and car (top-down view). Relative uses each frame's own camera as the origin; Absolute fixes the origin at the camera of $t_1$.}
    \label{fig:rel_abs_frames}
    \vspace{-44pt}
\end{wrapfigure}
Table~\ref{tab:qa_templates} lists the templates. \textit{[Prefix]} and \textit{[Camera]} are ``Following the camera,'' and ``the camera'' for relative questions, and ``Anchored to frame \{anchor\} of the clip,'' and ``the anchor camera'' for absolute spatial questions. Objects are identified by category and, when needed, an initial box in $[0,1000]$. Object-motion questions use separate paraphrases; their instructions specify the fixed anchor-camera frame. Table~\ref{tab:qa_answer_forms} gives geometric quantities and answer forms.

Figure~\ref{fig:rel_abs_frames} illustrates the two coordinate frames.

\paragraph{Reference-answer derivation.}
Table~\ref{tab:qa_construction_rules} defines the quantities and answer criteria in the question's reference frame. Answers combine one or two ordered phases; directions may combine axes (e.g., ``front-left'').

\paragraph{Filtering.}
Spatial QA retains tracks whose box area averages at least $1\%$ across frames and never exceeds $50\%$ of the image. A multimodal verifier rejects empty or background regions and clear label mismatches. We require valid target boxes and 3D centroids and readable clips. Two-target candidates require initial bounding-box IoU at most $0.5$. Selected windows within each video and QA collection have pairwise temporal IoU at most $0.5$.

\paragraph{Distractors.}
We favor the reference answer's phase count and exclude duplicate or equivalent options. For motion QA, we replace distractors consistent with the reference trajectory to retain one correct answer.

\paragraph{Missing-entity augmentation.}
We rewrite training questions to refer to objects verified by a multimodal model as absent throughout the clip. These objects have \texttt{bbox\_2d} and \texttt{centroid} set to \texttt{null} at each annotated timestep; the video and other states are unchanged. Template responses select an absence option. 

\begin{table}[!ht]
\centering
\footnotesize
\setlength{\tabcolsep}{4pt}
\renewcommand{\arraystretch}{1.0}
\begin{tabular}{@{}>{\raggedright\arraybackslash}p{0.21\linewidth}>{\raggedright\arraybackslash}p{0.12\linewidth}>{\raggedright\arraybackslash}p{0.61\linewidth}@{}}
\toprule
\textbf{Question type} & \textbf{Setting} & \textbf{Question template} \\
\midrule
Camera--object distance & Rel./Abs. & \textit{[Prefix]} how does the distance between \textit{[Object 1]} and \textit{[Camera]} change? \\
Object--object distance & Rel. & \textit{[Prefix]} how does the distance between \textit{[Object 1]} and \textit{[Object 2]} change? \\
Camera--object direction & Rel./Abs. & \textit{[Prefix]} how does the direction from \textit{[Camera]} to \textit{[Object 1]} change? \\
Object--object direction & Rel./Abs. & \textit{[Prefix]} how does the direction from \textit{[Object 1]} to \textit{[Object 2]} change? \\
Speed & Rel./Abs. & \textit{[Prefix]} how does the speed of \textit{[Object 1]} change? \\
Speed comparison & Rel./Abs. & \textit{[Prefix]} compare the speed between \textit{[Object 1]} and \textit{[Object 2]}. \\
Object motion & Abs. & How does \textit{[Object 1]} move over time? Answer in temporal order. \\
\bottomrule
\end{tabular}
\caption{Question templates for the 12 benchmark categories. Shared templates are grouped across coordinate settings. The object-motion row shows one of six paraphrases used in the data.}
\label{tab:qa_templates}
\end{table}

\begin{table}[!t]
\centering
\footnotesize
\setlength{\tabcolsep}{4pt}
\renewcommand{\arraystretch}{1.0}
\begin{tabular}{@{}>{\raggedright\arraybackslash}p{0.21\linewidth}>{\raggedright\arraybackslash}p{0.33\linewidth}>{\raggedright\arraybackslash}p{0.40\linewidth}@{}}
\toprule
\textbf{Question family} & \textbf{Reference quantity} & \textbf{Example choice text} \\
\midrule
Camera--object distance & Distance from the object to the current or anchor-camera origin. & It moves closer to the camera, then moves farther from the camera. \\
Object--object distance & Distance between two object centroids. & They move closer together, then move farther apart. \\
Direction & Directed 3D offset from the first entity to the second. & It is to the front, then to the front-right. \\
Speed & Magnitude of consecutive coordinate increments. & It slows down, then moves at a steady pace. \\
Speed comparison & Two objects' increment magnitudes over corresponding intervals. & The first is faster, then the second is faster. \\
Object motion & Translation along the fixed anchor camera's forward and lateral axes. & First moves toward the camera, then left \\
\bottomrule
\end{tabular}
\caption{Geometric quantities and example answer-option forms. Each option describes one or two ordered phases; the reference answer identifies one of the four options. The examples are drawn from the stored QA records.}
\label{tab:qa_answer_forms}
\end{table}

\begin{table}[!tb]
\centering
\footnotesize
\renewcommand{\arraystretch}{1.0}
\begin{tabular}{@{}>{\raggedright\arraybackslash}p{0.24\linewidth}@{\hspace{0.03\linewidth}}>{\raggedright\arraybackslash}p{0.73\linewidth}@{}}
\toprule
\textbf{Question type} & \textbf{How the answer is determined} \\
\midrule
\textbf{Distance}\newline
3D separation between the queried entities.
& \textbf{Nearly constant:} within $20\%$ of the phase's starting distance.\newline
\textbf{Increasing:} each distance exceeds the preceding one by more than $20\%$.\newline
\textbf{Decreasing:} each distance is more than $20\%$ below the preceding one. \\
\specialrule{0.3pt}{4pt}{5pt}
\textbf{Speed}\newline
3D displacement magnitude per interval.
& \textbf{Steady pace:} within $20\%$ of the phase's starting displacement.\newline
\textbf{Speeding up:} each displacement exceeds the preceding one by more than $20\%$.\newline
\textbf{Slowing down:} each displacement is more than $20\%$ below the preceding one. \\
\specialrule{0.3pt}{4pt}{5pt}
\textbf{Speed comparison}\newline
Two objects over the same interval.
& \textbf{Faster:} an object's displacement is more than $1.2$ times the other's.\newline
\textbf{Nearly equal:} neither object meets this condition. \\
\specialrule{0.3pt}{4pt}{5pt}
\textbf{Direction}\newline
Unit direction: first to second entity.
& \textbf{Front / left / above:} forward / left / up component above $0.3$.\newline
\textbf{Behind / right / below:} that component is below $-0.3$.\newline
Combine directions when multiple axes qualify. \\
\specialrule{0.3pt}{4pt}{5pt}
\textbf{Moving}\newline
Scene-normalized path in the anchor frame.
& \textbf{Stationary:} both forward--lateral and vertical path lengths are at most $0.04$.\newline
\textbf{Moving:} dominant forward, backward, left, or right displacement in each valid phase. \\
\bottomrule
\end{tabular}
\caption{Main criteria for determining reference answers. Distance, speed, speed comparison, and direction criteria are adapted from DSR-Suite~\citep{zhou2026learning} (Table~8 and implementation). Moving questions use our additional criteria for fixed-frame translation.}
\label{tab:qa_construction_rules}
\end{table}

\subsection{CoT Generation Settings}
\label{app:cot_generation_settings}
\label{app:cot_prompts}
\label{app:cot_validation}

\begin{figure}[!t]
    \centering
    \includegraphics[width=0.85\linewidth]{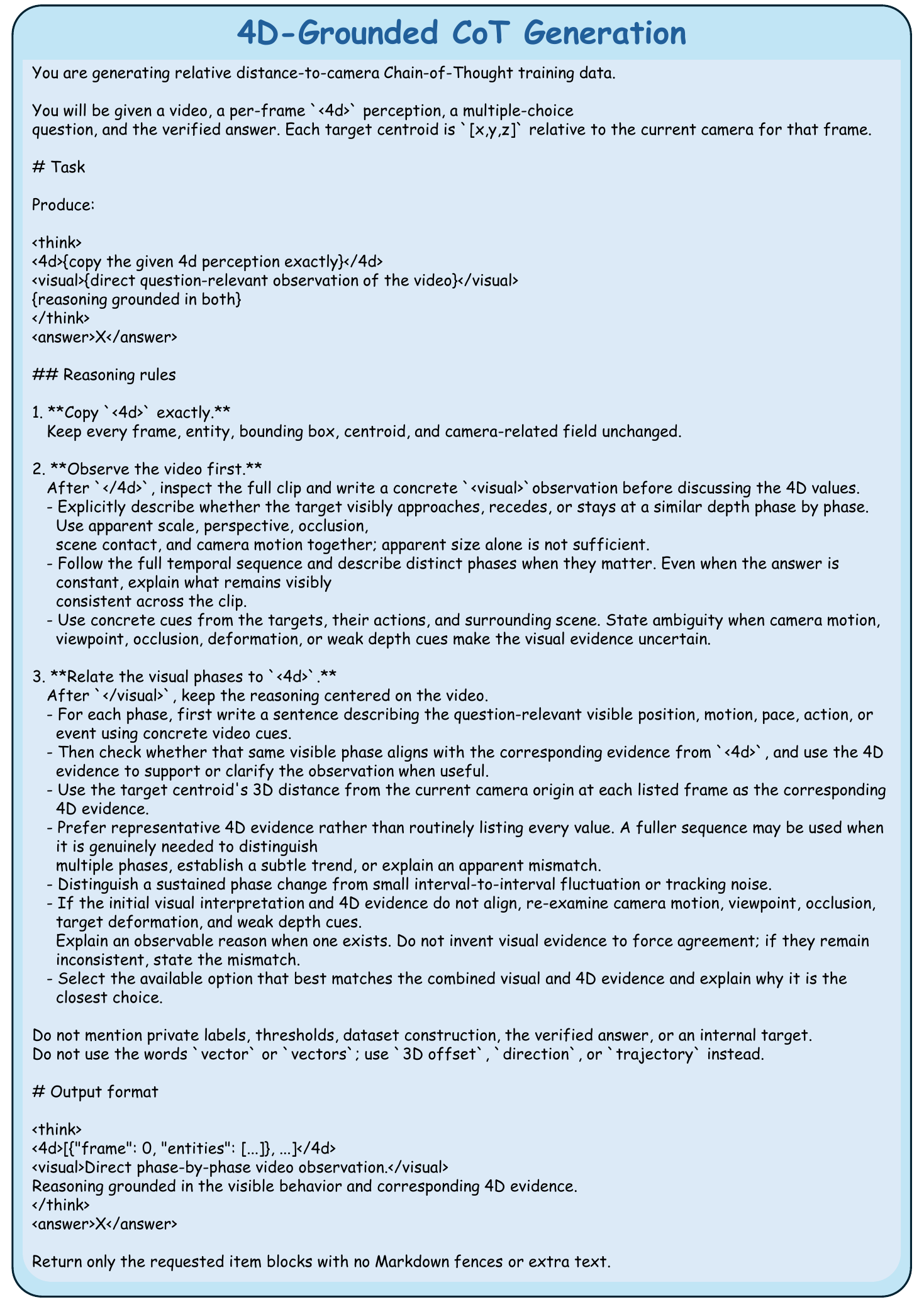}
    \caption{Teacher prompt for generating 4D-grounded CoT supervision, shown for relative distance-to-camera questions.}
    \label{fig:cot_generation_prompt}
\end{figure}

\paragraph{Teacher and inputs.}
GPT-5.6 (\texttt{gpt-5.6-sol}) receives the question, four options, reference states, correct answer, and 32 uniformly sampled frames. Frames are indexed, ordered temporally, and resized to fit $384\times216$ pixels while preserving aspect ratio. Task-specific prompts specify coordinate conventions and link visual observations to state changes (Figure~\ref{fig:cot_generation_prompt}).

\paragraph{Validation.}
We regenerate outputs from the same inputs if they alter reference states or the answer, lack a nonempty \texttt{<visual>} block, or violate the response format.

\clearpage

\section{Training and Evaluation Details}
\label{app:training_details}

\subsection{Optimization Settings}

Joint SFT trains for one epoch on Trace4D-Train: 600K 4D perception examples and approximately 70K grounded QA examples, mixed from the outset.

LoRA uses rank $r=128$, scaling factor $\alpha=256$, and dropout $0.05$. The language model and multimodal projector use cosine learning-rate decay from $1\mathrm{e}{-4}$ to $1\mathrm{e}{-5}$ with $3\%$ warmup. SFT runs on eight H100 GPUs with an effective batch size of 128.

\subsection{Ablation Settings}
\label{app:cot_ablation}
\label{app:stagewise_ablation}

Perception-first SFT uses one epoch of perception pretraining and one of QA adaptation. Adaptation uses all grounded QA examples and resamples $28\%$ of each relative and absolute perception pool per epoch for replay; this is the pool sampling rate, not the proportion in the training mixture.

All CoT variants, including Visual-only CoT, share this schedule without GRPO and differ only in QA supervision and instructions. Adaptation-only SFT runs the adaptation phase from the base VLM, whereas Joint SFT mixes perception and QA from the outset. Their budgets and data exposure differ. Table~\ref{tab:stagewise_ablation} also reports Joint SFT followed by GRPO.

\subsection{Geometry-Aware GRPO}
\label{app:grpo_details}

GRPO trains on 7,008 QA examples for one epoch with eight responses per input, cosine learning-rate decay from $1\mathrm{e}{-5}$, $3\%$ warmup, and KL coefficient $\beta=0.1$. In Equation~\ref{eq:composite_reward}, 4B uses $w_{\mathrm{acc}}=0.5$ and $w_{\mathrm{pos}}=w_{\Delta}=0.15$, whereas 8B uses $w_{\mathrm{acc}}=0.3$ and $w_{\mathrm{pos}}=w_{\Delta}=0.25$. Both use $w_{\mathrm{fmt}}=w_{\mathrm{bbox}}=0.1$.

The answer-accuracy reward checks the final option against $a_{\mathrm{ref}}$. The format reward requires valid 4D JSON and nonempty temporal visual analysis and 4D-grounded reasoning:
\begin{equation}
R_{\mathrm{acc}}(y)=
\begin{cases}
1, & \text{if the answer is correct},\\
0, & \text{otherwise},
\end{cases}
\qquad
R_{\mathrm{fmt}}(y)=
\begin{cases}
1, & \text{if }y\text{ is correctly formatted},\\
0, & \text{otherwise}.
\end{cases}
\label{eq:answer_format_rewards}
\end{equation}
For non-null object boxes, the IoU reward is
\begin{equation}
R_{\mathrm{bbox}}(y)
=\underset{i}{\operatorname{mean}}\!\left[
\underset{t}{\operatorname{mean}}\,
\operatorname{IoU}\!\left(\hat{\mathbf{b}}_{i,t},\bar{\mathbf{b}}_{i,t}\right)
\right],
\label{eq:bbox_reward}
\end{equation}
averaging over matched timesteps $t$ within each object track $i$, then across objects, excluding cameras. Matching uses exact entity labels and frame indices. Invalid predictions contribute zero to the corresponding geometric reward. Reference-null entries receive reward for matching absence. Displacements require valid matches at consecutive reference timesteps; entities without such intervals use position reward alone.

\begin{figure}[!t]
    \centering
    \begin{minipage}[t]{0.49\linewidth}
        \vspace{0pt}
        \includegraphics[width=\linewidth]{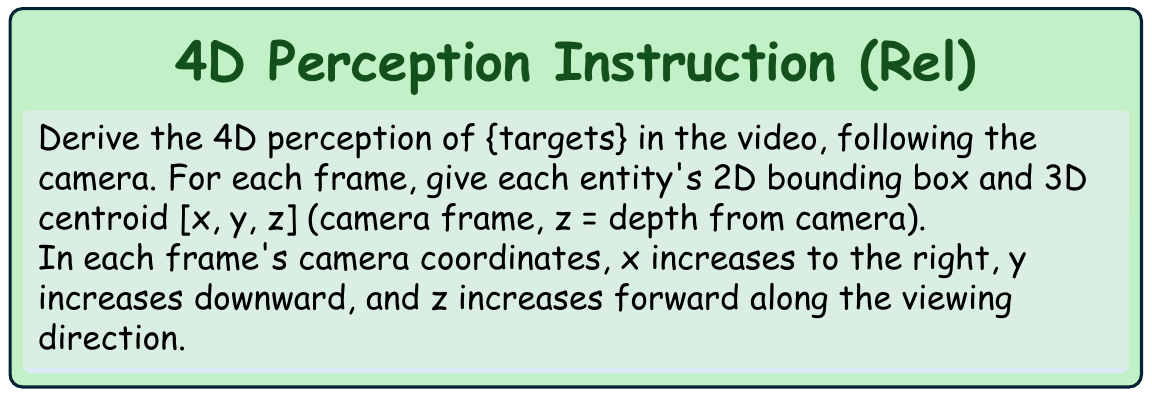}
    \end{minipage}\hfill
    \begin{minipage}[t]{0.49\linewidth}
        \vspace{0pt}
        \includegraphics[width=\linewidth]{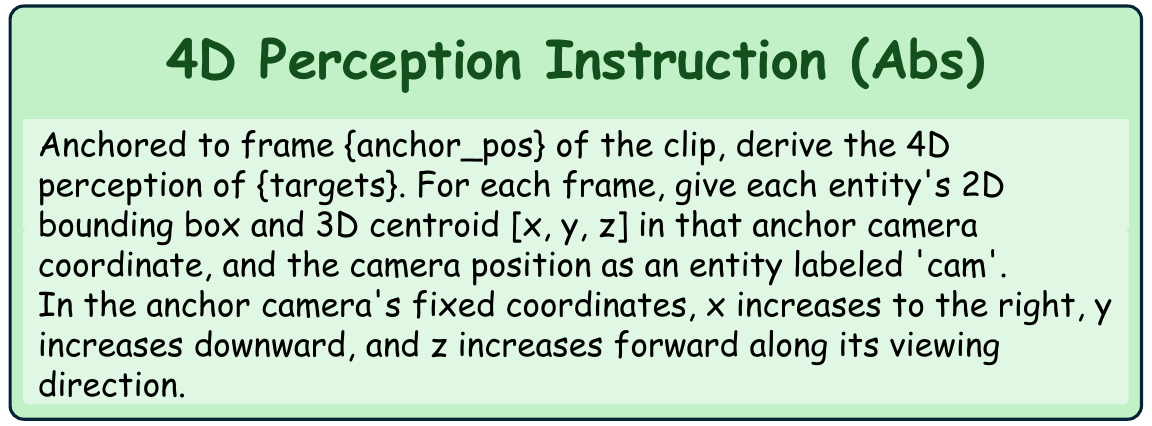}
    \end{minipage}
    \caption{Instructions for 4D perception supervision: relative (left) and absolute (right).}
    \label{fig:perception_training_suffixes}
\end{figure}

\begin{figure}[!t]
    \centering
    \begin{minipage}[t]{0.49\linewidth}
        \vspace{0pt}
        \includegraphics[width=\linewidth]{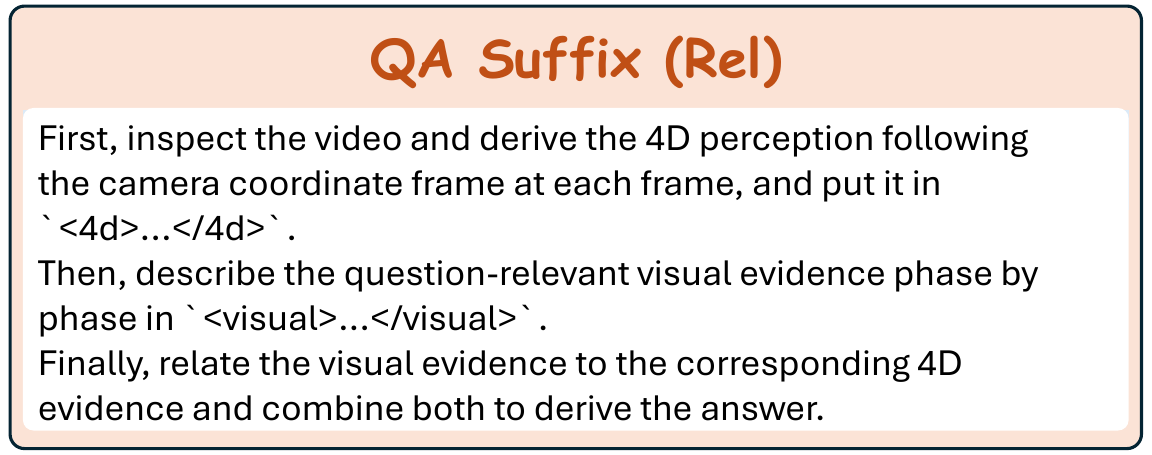}
    \end{minipage}\hfill
    \begin{minipage}[t]{0.49\linewidth}
        \vspace{0pt}
        \includegraphics[width=\linewidth]{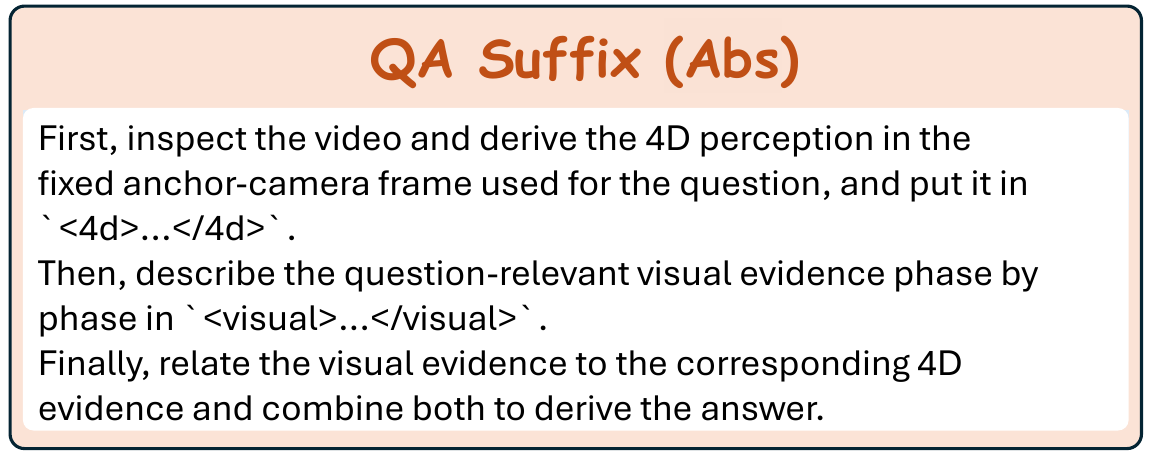}
    \end{minipage}
    \caption{QA suffixes for grounded QA training and inference: relative (left) and absolute (right).}
    \label{fig:qa_training_suffixes}
\end{figure}

\subsection{Training and Inference Prompts}
\label{app:shared_prompts}

Figure~\ref{fig:perception_training_suffixes} shows the relative and absolute instructions for state prediction; these examples contain no visual analysis or QA answer. Figure~\ref{fig:qa_training_suffixes} shows the suffixes appended to questions and options during grounded QA training and inference. They specify the coordinate convention and request the full 4D-grounded CoT and answer. Reference states and answer labels are training targets, not extra student inputs. The teacher prompt used to construct CoT supervision is shown in Figure~\ref{fig:cot_generation_prompt}.

\newpage
\subsection{Evaluation Protocol}
\label{app:evaluation_protocol}

\paragraph{QA evaluation.}
Trace4D uses 32 sampled video frames and greedy decoding with a maximum of 4,096 generated tokens. We extract the option letter from the final \texttt{<answer>} block; responses without a valid answer are counted as incorrect. Trace4D-Bench contains 1,200 questions over 1,178 distinct temporal clips, with 600 questions in each coordinate setting. VLM4D contains 1,816 questions over 1,000 videos (600 real and 400 synthetic). Overall accuracy is computed over all questions in each benchmark.

\paragraph{Perception metrics.}
We evaluate predicted states from 600 Relative and 600 Absolute spatial QA responses at exact matching frames (Tables~\ref{tab:perception_eval} and~\ref{tab:grpo_ablation}). IoU excludes supplied initial boxes; centroid and trajectory L2 measure position and consecutive-matched-frame displacement errors. We average within each valid object track, then across tracks, excluding cameras. For Relative, all reference coordinates are divided by the largest positive depth of the referenced objects at the first clip frame, shared across objects and timesteps. Absolute references are already normalized and evaluated as stored. Predictions are unchanged; no alignment is fitted.

\section{GRPO Ablation}
\label{app:grpo_ablation}

Table~\ref{tab:grpo_ablation} compares different reward configurations and their effects on QA accuracy and 4D state prediction.
\vspace{-8pt}
\begin{table}[H]
\centering
\footnotesize
\setlength{\tabcolsep}{4pt}
\renewcommand{\arraystretch}{1.12}
\begin{tabular*}{\linewidth}{@{\extracolsep{\fill}}lccccc@{}}
\toprule
\textbf{Training / feedback} & \multicolumn{2}{c}{\textbf{QA accuracy (\%)}} & \multicolumn{3}{c}{\textbf{4D state prediction}} \\
\cmidrule(lr){2-3}\cmidrule(lr){4-6}
 & \textbf{Overall} & \textbf{VLM4D} & \textbf{Centroid L2} $\downarrow$ & \textbf{Trajectory L2} $\downarrow$ & \textbf{IoU} $\uparrow$ \\
\midrule
Joint SFT & 77.83 & 62.39 & 0.1873 & 0.1178 & 0.7353 \\
+ GRPO: accuracy only & 78.33 & 62.67 & 0.1831 & 0.1164 & 0.7362 \\
+ GRPO: position only & 78.83 & 63.11 & 0.1805 & 0.1151 & \textbf{0.7403} \\
\textbf{Trace4D (Joint SFT + GRPO)} & \textbf{79.25} & \textbf{64.32} & \textbf{0.1775} & \textbf{0.1123} & 0.7377 \\
\bottomrule
\end{tabular*}
\caption{QA accuracy and 4D state prediction in the GRPO ablation. State metrics use all 1,200 Trace4D-Bench questions, with valid-track averaging.}
\label{tab:grpo_ablation}
\end{table}

\vspace{-8pt}

\paragraph{Settings.}

All variants use the same 7,008 QA examples and optimization settings (Appendix~\ref{app:grpo_details}). Accuracy-only sets $w_{\mathrm{acc}}=0.9$ and $w_{\mathrm{fmt}}=0.1$, with all geometric reward weights set to zero. Position-only and Trace4D share $w_{\mathrm{acc}}=0.5$ and $w_{\mathrm{fmt}}=w_{\mathrm{bbox}}=0.1$. Position-only uses $w_{\mathrm{pos}}=0.3$ and $w_{\Delta}=0$, while Trace4D uses $w_{\mathrm{pos}}=w_{\Delta}=0.15$.

\paragraph{State evaluation.}

We score 4D states in saved QA responses, averaging first within and then across object tracks. Cameras and initial-frame boxes are excluded. Models use their own matched timesteps.

\paragraph{Results.}

Trace4D outperforms Accuracy-only by 0.92 and 1.65 percentage points on Trace4D-Bench and VLM4D, respectively. Relative to Position-only, Trace4D also improves QA accuracy and reduces both centroid and trajectory L2 errors, though with slightly lower IoU. These results indicate that incorporating geometric rewards can improve QA accuracy while reducing coordinate and trajectory errors in the predicted 4D states.
\clearpage
\section{Qualitative Examples}
\label{app:reasoning_examples}

\begin{figure}[H]
    \centering
    \includegraphics[width=\linewidth,height=0.86\textheight,keepaspectratio]{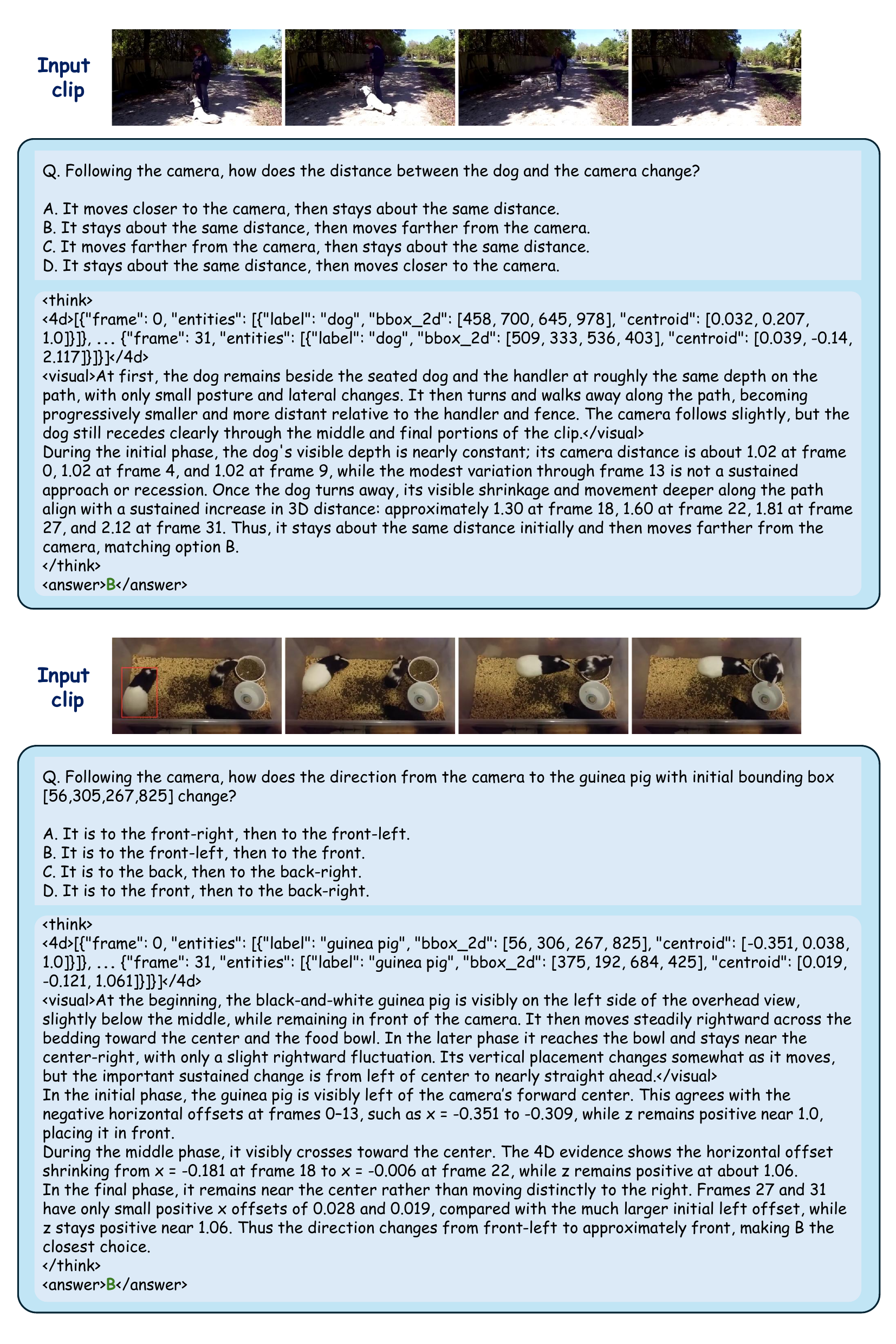}
    \caption{Trace4D predictions for camera-relative distance (top) and direction (bottom). Each example connects temporal visual observations with 4D state changes to support the answer. The displayed \texttt{<4d>} blocks are abbreviated.}
    \label{fig:reasoning_relative_examples}
    \label{fig:reasoning_distance_example}
    \label{fig:reasoning_direction_example}
\end{figure}

\begin{figure}[H]
    \centering
    \includegraphics[width=\linewidth,height=0.86\textheight,keepaspectratio]{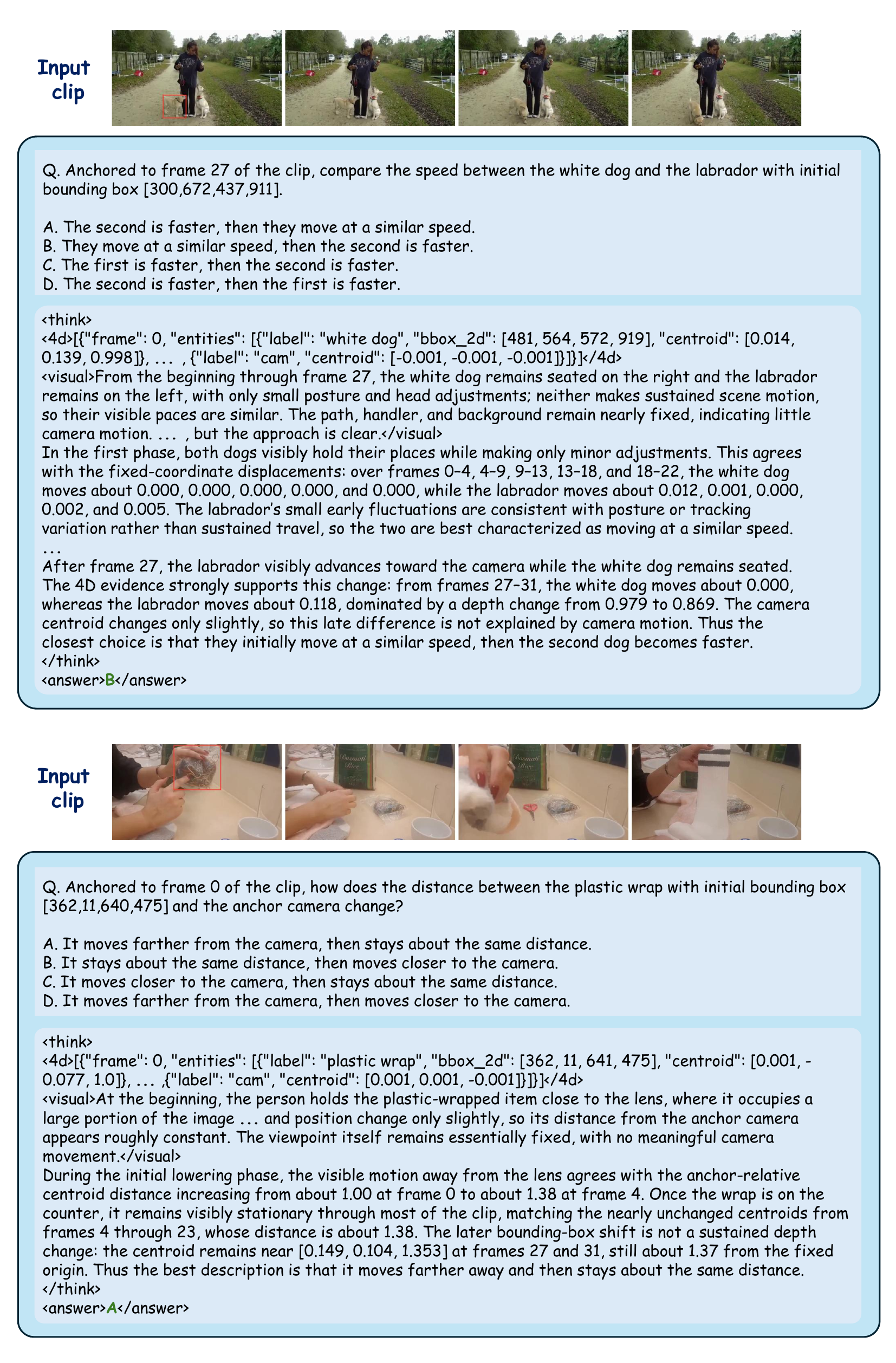}
    \caption{Trace4D predictions for absolute speed comparison (top) and distance to the anchor camera (bottom). Both examples use the fixed anchor-camera frame specified in the question. Ellipses indicate omitted portions of the responses.}
    \label{fig:reasoning_absolute_examples}
\end{figure}

\begin{figure}[H]
    \centering
    \includegraphics[width=\linewidth,height=0.86\textheight,keepaspectratio]{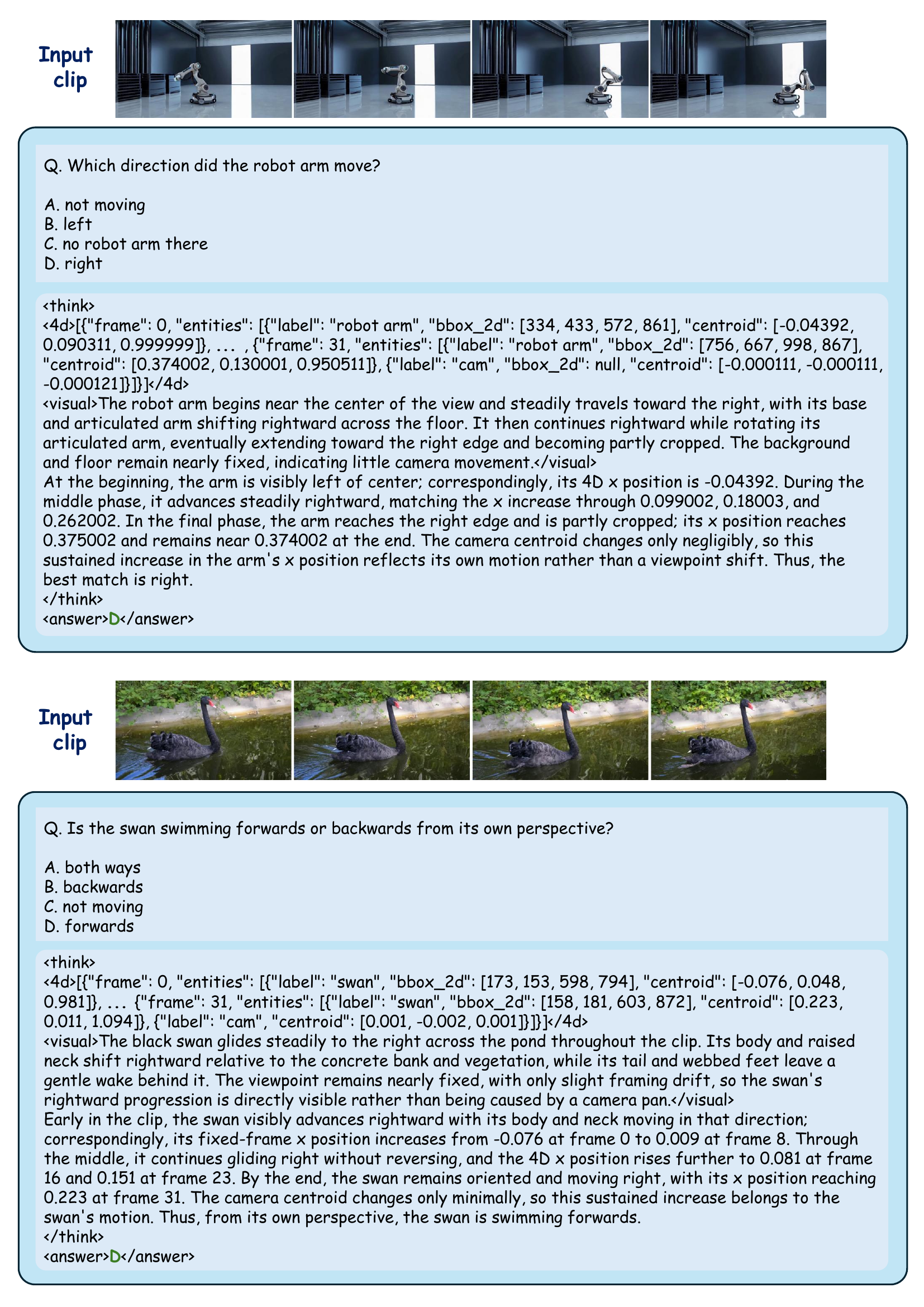}
    \caption{Trace4D predictions on VLM4D: robot-arm movement direction (top) and swan motion from its own perspective (bottom). The displayed \texttt{<4d>} blocks are abbreviated.}
    \label{fig:reasoning_vlm4d_examples}
\end{figure}

Figures~\ref{fig:reasoning_relative_examples}, \ref{fig:reasoning_absolute_examples}, and~\ref{fig:reasoning_vlm4d_examples} show Trace4D predictions on relative QA, absolute QA, and VLM4D.

\clearpage

\clearpage

\end{document}